\pdfoutput=1 
\documentclass{article}
\usepackage{siunitx}

\usepackage[utf8]{inputenc}
\usepackage[margin=0.75in]{geometry} %
\usepackage{authblk} %
\usepackage{lineno}

\usepackage[backend=biber,style=nature,sorting=none]{biblatex} %
\let\cite=\supercite

\usepackage{algorithm,amsmath}

\usepackage{hyperref} 
\usepackage{cleveref}
\usepackage{xcolor}

\hypersetup{
  colorlinks   = true, %
  urlcolor     = blue, %
  linkcolor    = blue, %
  citecolor   = blue %
}

\usepackage[noend]{algpseudocode}
\usepackage{graphicx}
\usepackage[labelformat=simple]{subcaption} 
\usepackage[hypcap=true]{caption}
\usepackage[T1]{fontenc}
\usepackage{lmodern}
\usepackage[labelfont=bf]{caption}

\title{Routing algorithms as tools for integrating social distancing with emergency evacuation}

\author[1,*]{Yi-Lin Tsai}
\author[2]{Chetanya Rastogi}
\author[1,3,4]{Peter K. Kitanidis}
\author[3,5,6]{Christopher B. Field}

\affil[1]{Department of Civil and Environmental Engineering, Stanford University, Stanford, CA, USA}
\affil[2]{Department of Computer Science, Stanford University, Stanford, CA, USA}
\affil[3]{Woods Institute for the Environment, Stanford University, Stanford, CA, USA}
\affil[4]{Institute for Computational and Mathematical Engineering, Stanford University, Stanford, CA, USA}
\affil[5]{Department of Biology, Stanford University, Stanford, CA, USA}
\affil[6]{Department of Earth System Science, Stanford University, Stanford, CA, USA}
\affil[*]{yilin2@stanford.edu}

\begin{document}

\date{}
\vspace{-0.5in}

\maketitle

\begin{abstract}
  One of the lessons from the COVID-19 pandemic is the importance of social distancing, even in challenging circumstances such as pre-hurricane evacuation. To explore the implications of integrating social distancing with evacuation operations, we describe this evacuation process as a Capacitated Vehicle Routing Problem (CVRP) and solve it using a DNN (Deep Neural Network)-based solution (Deep Reinforcement Learning) and a non-DNN solution (Sweep Algorithm). A central question is whether Deep Reinforcement Learning provides sufficient extra routing efficiency to accommodate increased social distancing in a time-constrained evacuation operation. We found that, in comparison to the Sweep Algorithm, Deep Reinforcement Learning can provide decision-makers with more efficient routing. However, the evacuation time saved by Deep Reinforcement Learning does not come close to compensating for the extra time required for social distancing, and its advantage disappears as the emergency vehicle capacity approaches the number of people per household.
\end{abstract}
 
\textbf{Keywords:} deep reinforcement learning, multi-hazard risk mitigation and management, compound disaster preparedness and response, human-centered AI, social distancing, COVID-19, pandemic 

\textbf{This paper has been accepted and published in Nature Scientific Reports: \url{https://www.nature.com/articles/s41598-021-98643-z\#article-info}}

\textbf{Cite this article:}
\textit{Tsai, YL., Rastogi, C., Kitanidis, P.K. et al. Routing algorithms as tools for integrating social distancing with emergency evacuation. Sci Rep 11, 19623 (2021). https://doi.org/10.1038/s41598-021-98643-z}

\clearpage

\section*{Introduction}
As of August 11, 2021, COVID-19 had caused over 204 million confirmed cases and over 4.3 million deaths around the world \cite{dong2020interactive}. However, floods, wildfires, earthquakes, and landslides do not take a break during a pandemic. We have to prepare for the challenges of responding to a catastrophe like a major hurricane (also known as typhoon or cyclone) while also managing the risk of spreading severe acute respiratory syndrome coronavirus 2 (SARS-CoV-2) \cite{phillips2020compound}. More generally, we must prepare for multiple overlapping disasters \cite{nandintsetseg2018cold, zscheischler2018future}. Furthermore, the complexity of evacuation operations and shelter management increases when a major disaster happens during a pandemic \cite{pei2020compound, shultz2020cascading, shultz2020mitigating, shultz2020superimposed, tripathy2021flood}. Social distancing is an indispensable containment measure against the spread of SARS-CoV-2 \cite{te2020effects, contreras2021challenges} and many other agents that could lead to pandemics. Maintaining social distancing during rescue operations is a logistical and time-management challenge. 

Although governments worldwide are ramping up COVID-19 vaccination, many challenges remain. Furthermore, integrating social distancing with emergency evacuation is relevant to planning for future disasters even after COVID-19 fades as a major risk. Looking beyond COVID-19, we need to be ready for the next pandemic, especially given the fact that increasing emergence and transmission of infectious diseases and pandemics, such as influenza, SARS-CoV-1, and SARS-CoV-2, have been associated with climate change for several decades \cite{jones2008global, shaman2013nino, flahault2016climate, beyer2021shifts}. One of the important lessons from COVID-19 is the need to anticipate future pandemics and take social distancing into account in compound-disaster preparedness and response.

Here, we focus on the interaction of social distancing and flood evacuation scheduling during a pandemic. As of April 8, 2021, tropical cyclones were the weather and climate disasters in the U.S. that led to the largest economic losses (\$1,011.3 billion, CPI-adjusted to 2021 prices), the highest number of deaths (6,593 people), and the highest average event cost (\$19.4 billion per event, in 2021 prices) from 1980 to 2021 \cite{NOAA2021}. 30.1 million people were affected and 3,134 people were killed worldwide by 65 flood events that overlapped with the COVID-19 pandemic in 2020 \cite{walton2020climate}. 

The Natural Hazard Mitigation Saves Study  \cite{mmc2017natural} found that every dollar invested by federal agencies in disaster mitigation saves society \$6 in post-disaster recovery costs, but spending on disaster recovery is almost nine times higher than on disaster prevention \cite{globalrisk2019}. The much better benefit to cost ratio of effective preparation than recovery is a strong argument for studying ways to improve evacuation planning. One possibility is that DNN-based methods can add substantial efficiency to emergency evacuation route planning.

Machine learning and deep neural networks (DNN) find diverse applications in Earth system science studies to build new data-driven models. Examples include identifying multi-hazard and extreme weather patterns, predicting river runoff in ungauged catchments, and precipitation nowcasting \cite{javidan2021evaluation, schlef2019atmospheric, reichstein2019deep, racah2017extremeweather, shi2017deep}. These state-of-the-art geoscientific tools help us better predict important phenomena. Increasingly, we can use deep neural networks to improve behavioral responses to these weather and climate disasters. In particular, machine learning and DNN-based techniques have the potential to enhance climate change mitigation and adaptation, from smart buildings and climate modeling to disaster management \cite{rolnick2019tackling}.

Deep reinforcement learning, combining traditional reinforcement learning and deep neural networks, has been widely applied to computer games, self-driving cars, the game of Go, and robotics \cite{schrittwieser2020mastering, silver2017mastering, mnih2016asynchronous, silver2016mastering, mnih2015human}. Applications of deep reinforcement learning to facilitate disaster responses are beginning to appear in the literature  \cite{zhou2021wireless, yan2020mobirescue, ghannad2020prioritizing, lee2020multi, niroui2019deep, tsai2019deep, ganapathi2018using}. In any disaster situation, the efficient delivery of assistance can contribute to limiting damages \cite{ukkusuri2008location, ozdamar2008greedy, yi2007dynamic, han2006global}. However, to the best of our knowledge, none of the existing research applies the state-of-the-art deep reinforcement learning to enhance the routing efficiency of disaster evacuation operations during a pandemic. 

This study aims to improve evacuation operations in compound events that involve both a pandemic and a well-forecast disaster like a major hurricane. Specifically, we investigate the role of social distancing in extending evacuation timelines and increasing the number of emergency vehicles required to evacuate a city, and we explore the potential to increase evacuation efficiency through using optimized vehicle routing based on a DNN-based method (deep reinforcement learning)\cite{kool2018attention} in comparison to a non-DNN method (sweep algorithm). Unlike the sweep algorithm, which requires pre-defined operational rules for picking up evacuees, deep reinforcement learning uses deep neural networks (DNN) to learn the best possible strategies and adaptively look for optimal routing. 

We chose New Orleans as the inspiration and case study, starting with its plans, population, and area, though not the specifics of its geography. New Orleans has a 72-hour hurricane evacuation timeline, which plans to pick up, from their homes, evacuees who signed up for the Special Needs Registry and transport them to the city-wide rescue center, the Smoothie King Center, during a 42-hour window before tropical storm winds reach the coast \cite{plan, schrilla2019evacuation} (see ``Methods''). 

Therefore, we can describe this pre-hurricane evacuation process as a Capacitated Vehicle Routing Problem (CVRP) where, for each neighborhood, there is one vehicle repeating the process of starting from a local rescue center, going to several houses to pick up people, and returning to the rescue center when its vehicle capacity is reached, until the vehicle picks up every resident on the special needs registry in the neighborhood (see ``Methods''). 

New Orleans' special needs registry is intended to address the extra challenges faced by a known population of elderly and disabled individuals. For instance, the elderly are usually more vulnerable in flood events, as documented for Hurricane Katrina and the 1953 flood in the Netherlands \cite{jonkman2009loss, jonkman2007loss, pollard1978north}. People with disabilities are less likely to move to a rescue center by themselves and thus need more assistance from rescue teams during a flood event.

The COVID-19 pandemic adds significantly to the complexities of evacuation planning and execution. Craig Fugate, a former Federal Emergency Management Agency (FEMA) Administrator, said\cite{rott_2020}, ``We've been telling people: stay home, stay home, stay home, stay home. And then you're going to turn around and tell them they need to evacuate. That's going to be a hard message.'' Also, according to FEMA's COVID-19 Supplement for Planning Considerations: Evacuation and Shelter-in-Place, one of the critical questions for state, local, tribal, and territorial governments is ``Have you incorporated social distancing considerations when calculating evacuation clearance time (e.g., reduced load capacity, additional vehicles, increased loading time)? \cite{femacovid19}'' However, to the best of our knowledge, there is no previous research that aims to help jurisdictions answer this question. 

Although New Orleans asks evacuees to include face coverings in their ``go-bag''  \cite{plan}, social distancing is still important.  But social distancing, implemented through decreasing the number of people per rescue vehicle, adds to the time and distance vehicles need to cover. To assess the feasibility of social distancing as part of an evacuation operation, we explore: (1) total time of picking up and transporting evacuees from their houses to a rescue center; (2) sizes of neighborhood that one emergency vehicle serves for evacuation operations; (3) degrees of social distancing enforced in an emergency vehicle; and (4) the number of emergency vehicles needed to evacuate the whole city.

\section*{Results}

\subsection*{Trade-offs in size of neighborhood and social distancing}
In this section, we analyzed the trade-offs in size of neighborhood and social distancing by simulating the CVRP process of an emergency vehicle repeatedly starting from a rescue center to collect residents house by house until its vehicle capacity is met. In this study, we used the locations of nodes (or ``houses'') and a depot (a rescue center) in the standard CVRP benchmarking datasets published by \cite{cvrplib, augerat1995computational}. We generated the demand of each node (household size or the number of people in each house to be picked up by the rescue vehicle) using the average household size in New Orleans (see ``Methods''). The sizes of neighborhood in the four datasets we used in this study were 20, 35, 52, and 68 houses on the special needs registry. Social distancing limited the number of passengers allowed in one rescue vehicle, with capacities of 64, 32, 16, 8, 4, and 2 passengers per rescue vehicle. One can think of this range of vehicle capacities as representing the span from no social distancing to very strict social distancing in a large bus.  But it can also be a way to explore modest distancing (2- to 4-fold capacity changes) in vehicles ranging from a large bus to a small van or sedan.

\begin{figure}[H]\captionsetup{singlelinecheck = false, justification=justified}
	\centering
	\begin{subfigure}{0.48\textwidth} %
	    \caption{DNN-based solution} %
		\includegraphics[width=\textwidth]{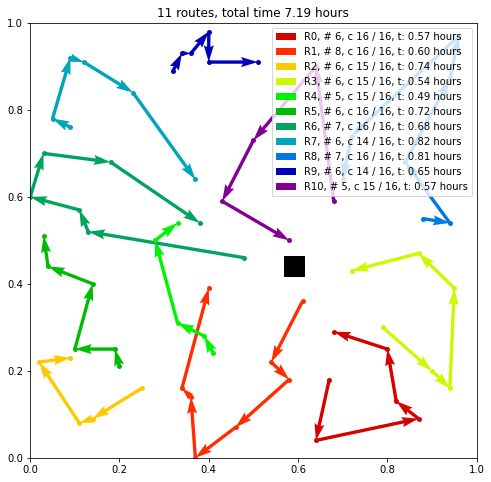}
	\end{subfigure}
	\vspace{0.1cm} %
	\begin{subfigure}{0.48\textwidth} %
	    \caption{Non-DNN solution} %
		\includegraphics[width=\textwidth]{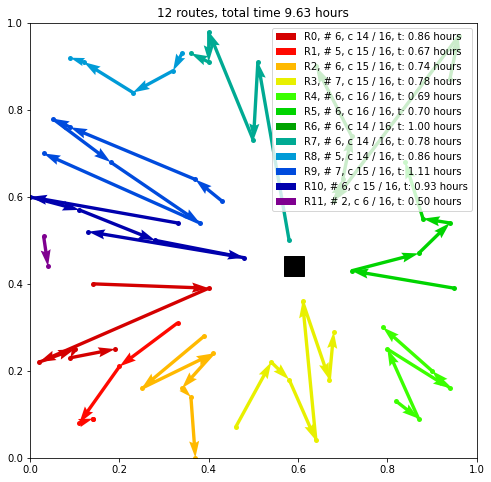}
	\end{subfigure}
	\caption{\textbf{An example of DNN-based and non-DNN solutions.} Information in each legend box is in the format \cite{kool2018attention} ``R0, \#6, c 14/16, t: 0.86 hours''. R0 is the first route, R1 is the second, etc. ``\#6'' means that the emergency vehicle visits 6 houses in that route. ``c 14/16'' indicates that the emergency vehicle whose passenger capacity is 16 people picks up 14 people in route R0. Finally, ``t: 0.86 hours'' means that the emergency vehicle spends 0.86 hours in route R0. To make the routing pattern easy to read \cite{kool2018attention}, we hide the first and the last line going from and back to the black square, which represents the transition point for transit to the depot. The total time and the number of routes required to pick up every resident on the special needs registry are shown at the top of Fig. \ref{fig:1}. This example is for vehicle capacity = 16 and transit time = 0.}\label{fig:1}.  %
\end{figure}

After the emergency vehicle picked up every resident on the special needs registry in the neighborhood, we summed the total time and the number of routes as the outputs (see Supplementary information for more information about the outputs). Figure \ref{fig:1} is an example of the output.

\begin{figure}[H]
\includegraphics[width=1\textwidth]{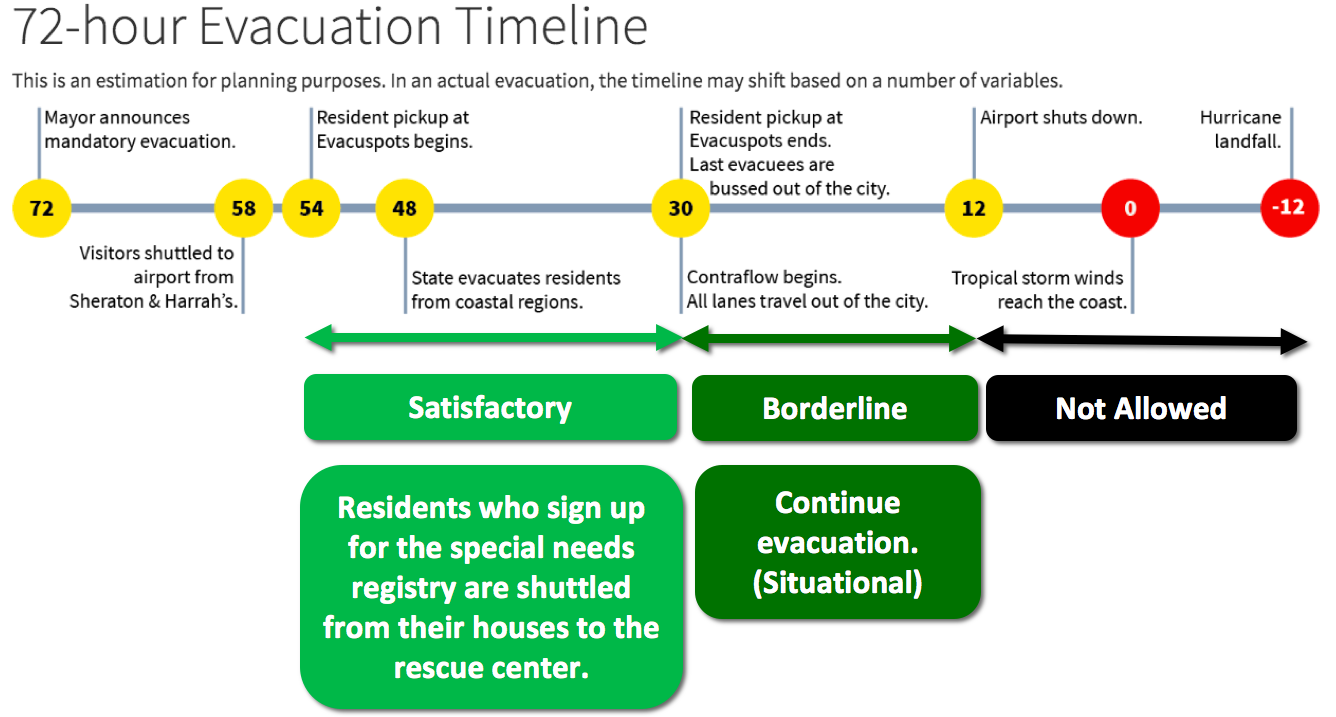}\centering
\caption{\textbf{72-hour evacuation timeline in New Orleans.} This pre-hurricane evacuation timeline is adapted from \cite{plan, schrilla2019evacuation}.}.\label{fig:timeline}
\centering
\end{figure}

New Orleans plans to pick up residents starting 54 hours before the storm reaches the coast and collect the last residents by 30 hours before the storm reaches the coast, for a 24-hour window (Fig. \ref{fig:timeline}) \cite{plan, schrilla2019evacuation}. Between 30 hours and 12 hours before the storm reaches the coast, the city may continue the evacuation operation if necessary \cite{schrilla2019evacuation}. Therefore, to evaluate the performance of the DNN-based and non-DNN models, we used 42 hours and 24 hours as the thresholds to determine if the vehicle completed the evacuation missions within the desired timelines. Further, we classified the time performance as Satisfactory (<24 hours), Borderline (24-42 hours), and Not Allowed (>42 hours). 

To accommodate the transit time required to travel from each neighborhood to the centrally located Smoothie King Center and back, we added 0, 0.5, 1, or 2 hours to each evacuation route. In a realistic city layout, transit times would differ among neighborhoods, but they should be  similar for each route within a neighborhood. With zero transit time, the de facto assumption is that the Smoothie King Center is in each neighborhood.  With additional fixed transit time per route, the black squares in Fig. \ref{fig:1} and other figures in Supplementary information represent the closest freeway on-ramp or intersection with a major thoroughfare, marking the transition from collecting people within a neighborhood to transporting people to the Smoothie King Center.

\subsubsection*{Size of neighborhood}
\begin{figure}[H]\captionsetup{singlelinecheck = false, justification=justified}
	\centering
	\begin{subfigure}{0.48\textwidth} %
	    \caption{DNN-based solution} %
		\includegraphics[width=\textwidth]{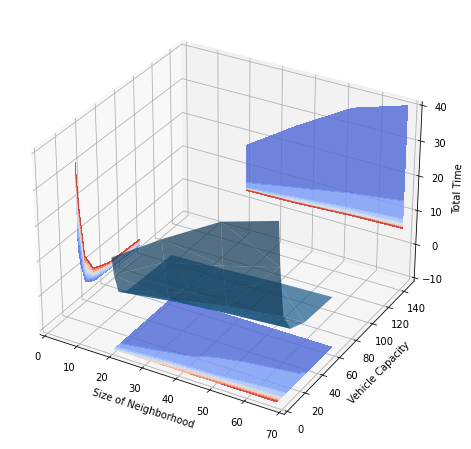}
	\end{subfigure}
	\vspace{0.1cm} %
	\begin{subfigure}{0.48\textwidth} %
		\caption{Non-DNN solution} %
		\includegraphics[width=\textwidth]{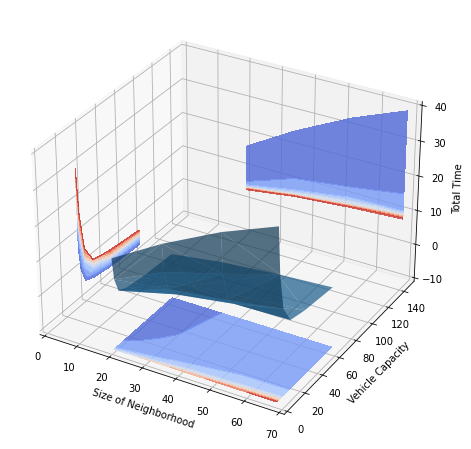}
	\end{subfigure}
	\caption{\textbf{Impacts of neighborhood size and vehicle capacity on total time.} (x axis: size of neighborhood (unit: the number of houses with people on the special needs registry), y axis: vehicle capacity due to social distancing (unit: people per vehicle), z axis: total time (unit: hours)).}\label{fig:3D_view} %
\end{figure}

With any level of capacity per rescue vehicle (social distancing), the total time for evacuations increased as the size of neighborhood (number of people on the special needs registry) increased (Fig. \ref{fig:3D_view}). The total time of both DNN-based and non-DNN solutions rose more steeply with stronger social distancing across all sizes of neighborhood. For example, with 32 people in an emergency vehicle instead of 64, increasing the neighborhood size from 20 houses to 68 (a factor of 3.4), increased the total time by 1.89 hours and 4.79 hours for DNN-based and non-DNN solutions respectively. However, with more stringent social distancing (2 people per vehicle), increasing the neighborhood size from 20 houses to 68 increased the total time by 23.37 hours and 22.18 hours for DNN-based and non-DNN solutions respectively.

\begin{figure}[H]
\includegraphics[width=0.5\textwidth]{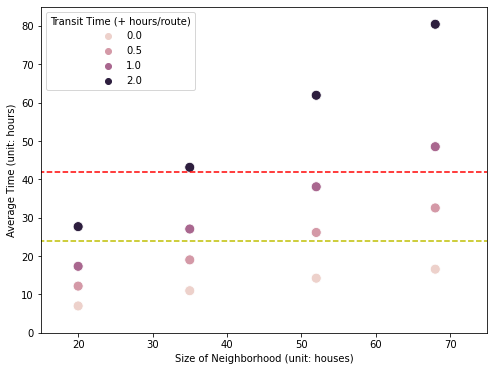}\centering
\caption{\textbf{Trade-offs between size of neighborhood and average evacuation time.} The y-axis is the average time across both algorithms and all vehicle capacities (passenger limit= 2, 4, 8, 16, 32, and 64 people per emergency vehicle). The red dashed line represents the 42-hour window (54-12 hours before tropical storm winds reach the coast) while the yellow dashed line represents the 24-hour window (54-30 hours before tropical storm winds reach the coast).}\label{fig:neighborhood_vs_time}
\centering
\end{figure}

The average total time across six vehicle capacities (2, 4, 8, 16, 32, and 64 people per emergency vehicle) and two algorithms (deep reinforcement learning and sweep algorithm) rises with the neighborhood size, and the change is linearly proportional to the change in neighborhood size (Fig. \ref{fig:neighborhood_vs_time}). For example, with 16 people per emergency vehicle, increasing the neighborhood size from 20 houses to 68 (a factor of 3.4), increased the total time by 2.37-fold and 2.97-fold for cases with the minimum (+0 hour/route) and the maximum (+2 hours/route) transit time respectively. Compared to the cases without adding any transit time to each route, the relationship between neighborhood size and the average total time is closer to linearly proportional in the cases with longer transit time adding to each route.

Without adding any transit time to each route, a disaster manager can evacuate everyone on the special needs registry in all sizes of neighborhood within 24 hours (achieving the Satisfactory threshold) (Fig. \ref{fig:neighborhood_vs_time}). In addition, the smaller the neighborhood size, the more likely every resident on the special needs registry can be evacuated within 24 hours (Satisfactory threshold) or 42 hours (Borderline threshold). With a transit time of two hours per route, only the smallest neighborhood can be fully evacuated by the end of 42 hours. Of course, the trade-off is that dividing a city into smaller neighborhoods increases the number of neighborhoods to evacuate.

\subsubsection*{Social distancing}
\begin{figure}[H]\captionsetup{singlelinecheck = false, justification=justified}
	\centering
	\begin{subfigure}{0.48\textwidth} %
		\caption{Neighborhood 1} %
		\includegraphics[width=\textwidth]{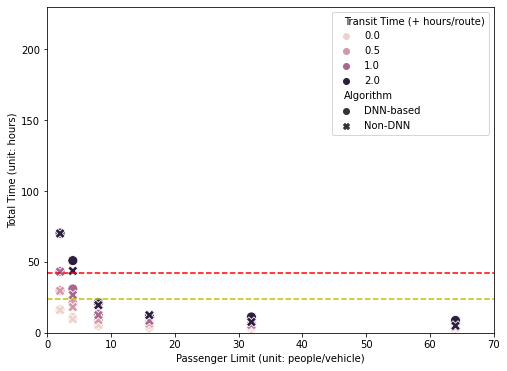}
	\end{subfigure}
	\vspace{0.1cm} %
	\begin{subfigure}{0.48\textwidth} %
		\caption{Neighborhood 2} %
		\includegraphics[width=\textwidth]{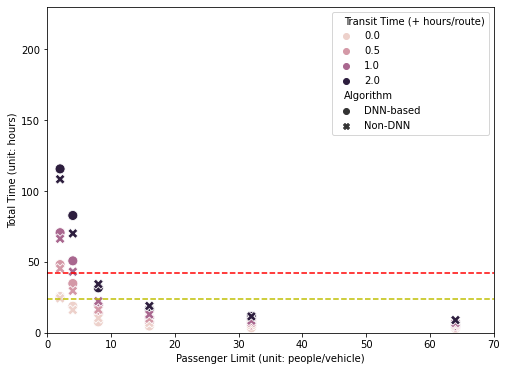}
	\end{subfigure}

	\begin{subfigure}{0.48\textwidth} %
		\caption{Neighborhood 3} %
		\includegraphics[width=\textwidth]{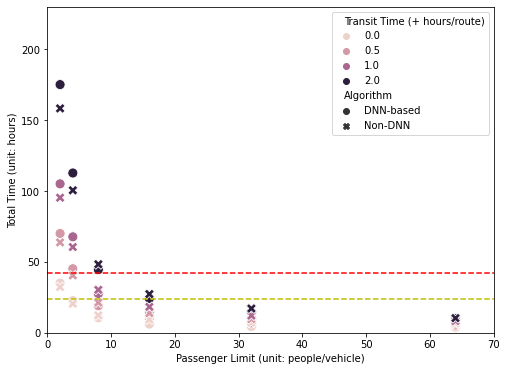}
	\end{subfigure}
	\vspace{0.1cm} %
	\begin{subfigure}{0.48\textwidth} %
		\caption{Neighborhood 4} %
		\includegraphics[width=\textwidth]{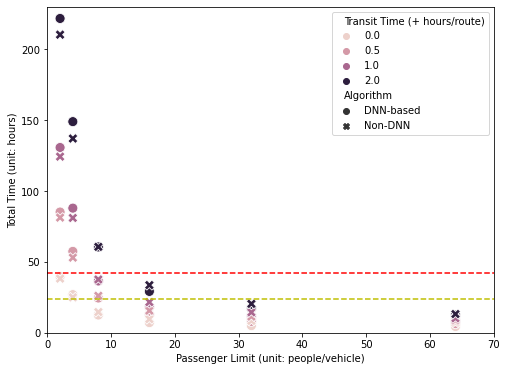}
	\end{subfigure}
	\caption{\textbf{Social distancing and total evacuation time for a neighborhood.} The x-axis indicates different social distancing protocols (the number of people allowed in an emergency vehicle) and y-axis indicates the total time required to evacuate a neighborhood. Red and yellow dashed lines indicate the 42-hour and 24-hour windows, respectively (Fig. \ref{fig:timeline})). \textbf{a} Neighborhood 1 has 52 people and 20 houses on the special needs registry. \textbf{b} Neighborhood 2 has 83 people and 35 houses. \textbf{c} Neighborhood 3 has 126 people and 52 houses. \textbf{d} Neighborhood 4 has 168 people and 68 houses.}\label{fig:social_distancing_vs_time}

\end{figure}

As expected, for any neighborhood size, algorithm, and transit time, stricter social distancing increases the total evacuation time (Fig. \ref{fig:social_distancing_vs_time}). For the strictest social distancing (2 people per vehicle), the evacuation time is less than 42 hours only when transit time is zero. With milder social distancing (32 people per vehicle), the sensitivity of evacuation time to passenger limit is approximately the same in the DNN-based and non-DNN solutions. For example, without any additional transit time per evacuation route, decreasing vehicle capacity from 64 to 32 increases evacuation time 1.13-1.17-fold and 1.12-1.23-fold in DNN-based- and non-DNN solution respectively. However, with stricter social distancing, evacuation times for the DNN-based solutions are more sensitive to vehicle capacity than are the non-DNN solutions (Fig. \ref{fig:social_distancing_vs_time}). Across the full range of neighborhood sizes, decreasing vehicle capacity from 64 to 2 increases evacuation time by up to 8.94-fold with a DNN-based solution and 5.41-fold in a non-DNN solution.

\begin{figure}[H]\captionsetup{singlelinecheck = false, justification=justified}
	\centering
	\begin{subfigure}{0.48\textwidth} %
		\caption{Neighborhood 1} %
		\includegraphics[width=\textwidth]{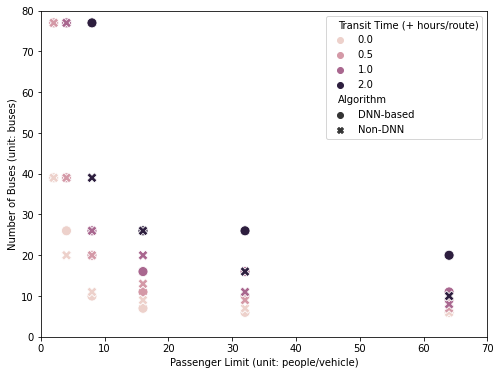}
	\end{subfigure}
	\vspace{0.1cm} %
	\begin{subfigure}{0.48\textwidth} %
		\caption{Neighborhood 2} %
		\includegraphics[width=\textwidth]{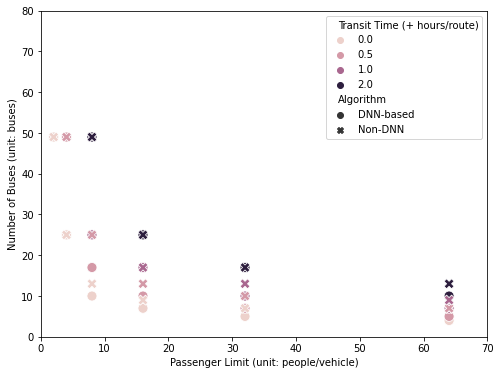}
	\end{subfigure}

	\begin{subfigure}{0.48\textwidth} %
		\caption{Neighborhood 3} %
		\includegraphics[width=\textwidth]{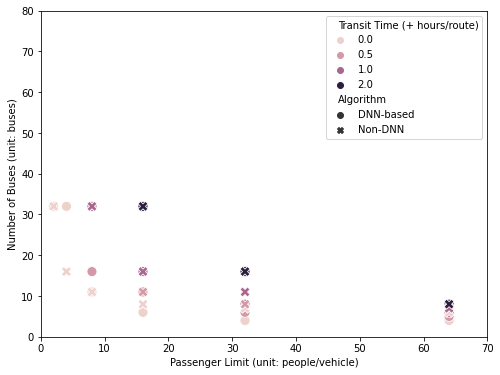}
	\end{subfigure}
	\vspace{0.1cm} %
	\begin{subfigure}{0.48\textwidth} %
		\caption{Neighborhood 4} %
		\includegraphics[width=\textwidth]{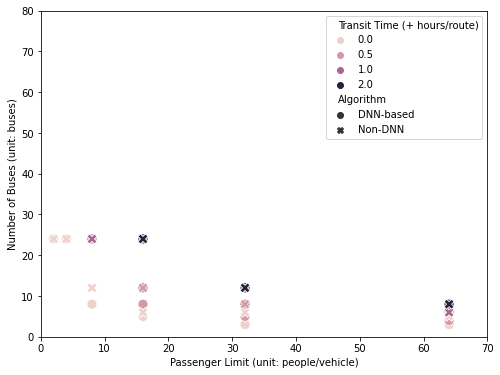}
	\end{subfigure}
	\caption{\textbf{Social distancing and the number of vehicles required to evacuate a city.} The x-axis is vehicle capacity, and the y-axis indicates the number of vehicles required to evacuate 4,000 people. If it is impossible to evacuate one neighborhood within 42 hours, data points are not shown. \textbf{a} Neighborhood 1 has 52 people and 20 houses on the special needs registry. \textbf{b} Neighborhood 2 has 83 people and 35 houses. \textbf{c} Neighborhood 3 has 126 people and 52 houses. \textbf{d} Neighborhood 4 has 168 people and 68 houses.}\label{fig:social_distancing_vs_bus}

\end{figure}

One of the critical elements of an evacuation plan is the number of vehicles to allocate. A rough estimate for this can come from the ratio of the size of the special needs registry to the number of people evacuated by a single vehicle within the 42-hour target window (Fig. \ref{fig:social_distancing_vs_bus}). With stricter social distancing, the number of emergency vehicles increased. Stricter social distancing also increased the risk that one emergency vehicle could not fully evacuate one neighborhood within 24 or 42 hours (Figs. \ref{fig:social_distancing_vs_time} and \ref{fig:social_distancing_vs_bus}).

\subsection*{Efficacy of DNN-based and non-DNN solutions}

\begin{figure}[H]\captionsetup{singlelinecheck = false, justification=justified}
	\centering
	\begin{subfigure}{0.48\textwidth} %
		\caption{Neighborhood 1} %
		\includegraphics[width=\textwidth]{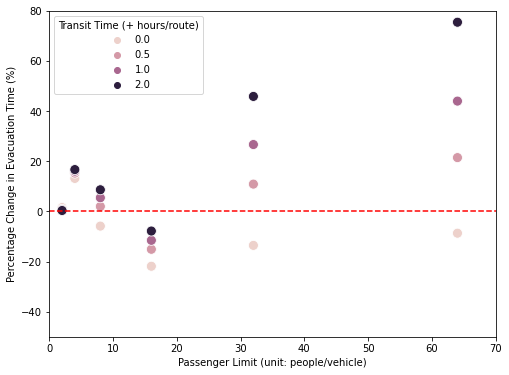}
	\end{subfigure}
	\vspace{0.1cm} %
	\begin{subfigure}{0.48\textwidth} %
		\caption{Neighborhood 2} %
		\includegraphics[width=\textwidth]{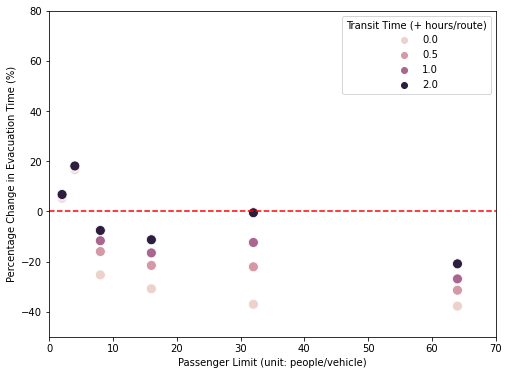}
	\end{subfigure}

	\begin{subfigure}{0.48\textwidth} %
		\caption{Neighborhood 3} %
		\includegraphics[width=\textwidth]{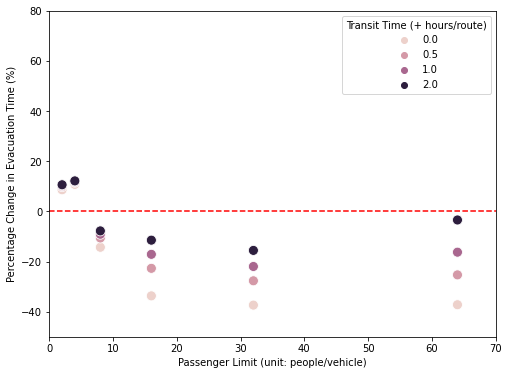}
	\end{subfigure}
	\vspace{0.1cm} %
	\begin{subfigure}{0.48\textwidth} %
		\caption{Neighborhood 4} %
		\includegraphics[width=\textwidth]{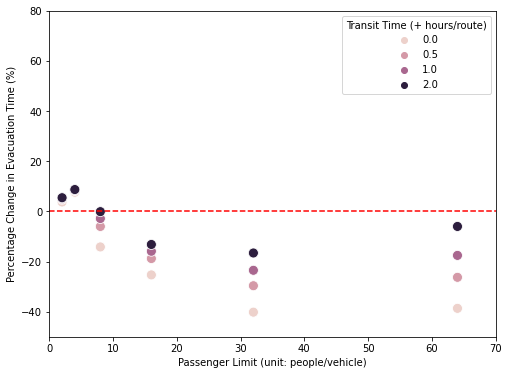}
	\end{subfigure}
	\caption{\textbf{Performance comparison between DNN-based and non-DNN solutions in terms of total evacuation time for one neighborhood.} The x-axis is vehicle capacity and the y-axis is the percentage change in evacuation time is calculated as equation (\ref{eq:1}). Points below the red dashed line indicate that the DNN-based solution saves time. \textbf{a} Neighborhood 1 has 52 people and 20 houses on the special needs registry. \textbf{b} Neighborhood 2 has 83 people and 35 houses. \textbf{c} Neighborhood 3 has 126 people and 52 houses. \textbf{d} Neighborhood 4 has 168 people and 68 houses.}\label{fig:time_comparison}
\end{figure}

Across all of the scenarios without additional transit time in each route, the DNN-based solution generated a shorter total evacuation time than the non-DNN solution in 66.67\% of the cases (Fig. \ref{fig:time_comparison}). The DNN-based solution required fewer routes than the non-DNN solution in 8.33\% of the scenarios. On average, the DNN-based method took less time but used more routes. With additional transit time in each route, the advantage of the DNN-based algorithm decreased (Fig. \ref{fig:time_comparison}). In general, the DNN-based solutions outperformed the non-DNN solutions, except when neighborhood size was very small (Fig. \ref{fig:time_comparison}a) or vehicle capacity was very low (Fig. \ref{fig:time_comparison}a-d). The DNN-based solutions that took less time were the ones that needed fewer or the same number of routes. When DNN-based solutions took more time, they required more routes.

\begin{equation} \label{eq:1} {Percentage\ Change\ in\ Evacuation\ Time\ (\%)}= \frac{Time_{DNN-based}-Time_{Non-DNN}}{Time_{Non-DNN}}\times 100 \\
\end{equation}

Overall, 63.54\% of the DNN-based solutions met the threshold for evacuation within 24 hours, see Fig. \ref{fig:timeline}), the same as the non-DNN solutions (Fig. \ref{fig:social_distancing_vs_time}). Times were greater than 42 hours in 22.92\% and 21.88\% of the DNN-based and non-DNN solutions respectively. In 1.04\% of the cases, the non-DNN solution was enough better than the DNN-based solution to shift the time performance from > 42 hours to between 24 and 42 hours. Even though the DNN-based solutions used less time than the non-DNN solutions in 57.29\% of the cases, none of the DNN-based solutions were enough better than the non-DNN solutions to shift a result from > 42 to 24 to 42 or from 24 to 42 to < 24.

DNN-based solutions outperformed non-DNN solutions by up to 40.18\% (Fig. \ref{fig:time_comparison}). The advantage of the DNN-based approach was largest with high-capacity vehicles, mild social distancing, and larger neighborhoods (Fig. \ref{fig:time_comparison}b-d). In visual terms, the DNN-based method outperformed the non-DNN method because its routing patterns were smoother loops compared to the non-DNN method, which wasted a lot of time going back and forth among houses with its more ``spiky'' routing (Fig. \ref{fig:1} and the other figures in Supplementary information).

However, with the non-DNN approach, the vehicle always picked up passengers until it reached its capacity, while in some of the DNN-based solutions, the vehicle returned to the rescue center before reaching full capacity (see Supplementary Fig. S1). This led to the DNN-based solutions sometimes using more routes than the non-DNN solutions. An added transit time was effectively a penalty on the number of routes per neighborhood, leading to DNN-based solutions using up to 75.38\% more time than the non-DNN solutions (Fig. \ref{fig:time_comparison}a).

\begin{figure}[H]\captionsetup{singlelinecheck = false, justification=justified}
	\centering
	\begin{subfigure}{0.45\textwidth} %
		\caption{Comparison of the number of routes} %
		\includegraphics[width=\textwidth]{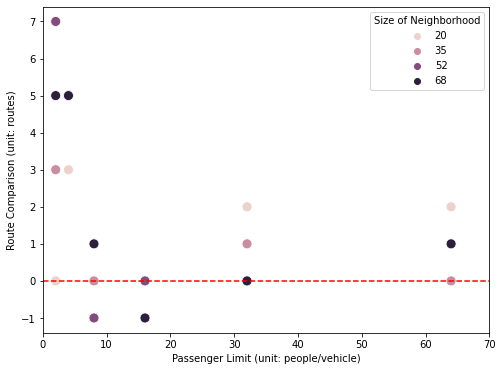}
	\end{subfigure}
	\vspace{0.1 cm} %
	\begin{subfigure}{0.45\textwidth} %
		\caption{Percentage change in the number of routes} %
		\includegraphics[width=\textwidth]{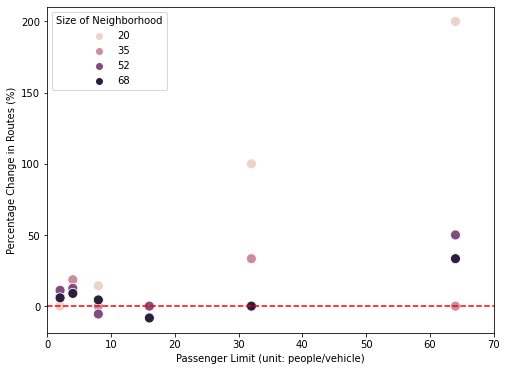}
	\end{subfigure}
	\caption{\textbf{Performance comparison between DNN-based and Non-DNN solutions in terms of the number of routes.} Negative numbers on the y-axis means that the DNN-based solution required fewer routes than the non-DNN solution. The x-axis is vehicle capacity and the y-axis is the difference in the number of routes (Fig. \ref{fig:route_comparison}a) and the percentage change in the number of routes (Fig. \ref{fig:route_comparison}b, equation (\ref{eq:2})) is calculated as follows.}\label{fig:route_comparison} %
\end{figure}

The number of routes required for DNN-based and non-DNN solutions were approximately the same most of the time (Fig. \ref{fig:route_comparison}). In 58.33\% of the cases the difference is within ${\pm}10{\%}$. In 33.33\% of the cases, DNN-based and non-DNN solutions used the same number of routes (Fig. \ref{fig:route_comparison}b). In these cases, the transit time per route did not impact total evacuation time too much, and the advantage of the DNN-based solutions was expressed (Fig. \ref{fig:time_comparison}b-d).

\begin{equation} \label{eq:2}{Percentage\ Change\ in\ the\ Number\ of\ Routes\ (\%)}= \frac{Routes_{DNN-based}-Routes_{Non-DNN}}{Routes_{Non-DNN}}\times 100 \\
\end{equation}

The advantage of deep reinforcement learning faded with smaller vehicles or stricter social distancing. When the vehicle capacity was close to the number of people in each household, the DNN-based solution performed about the same as or even worse than the sweep algorithm. It is not surprising that the DNN-based method did not work well when vehicle capacity was close to the household size. This is a situation known as the demand of a single node in the Capacitated Vehicle Routing Problem. If the vehicle collects two people from one house and then returns to the rescue center, the problem collapses to one of determining the shortest path between each node and the depot (Supplementary Fig. S6). The order of visiting houses might be different, but no version of a DNN-based approach would outperform a non-DNN approach.

We did not specifically train the DNN-based model for vehicle sizes close to the number of people in a house, largely because the fact that all solutions converge at some point indicates that this retraining could provide no more than marginal improvements. In addition, a DNN-based model is more complicated than is appropriate for these simple edge cases.

Across all scenarios, the DNN-based method required 1.67 more routes (20.96\%) than the non-DNN solution but saved 0.73 hours (14.78\%), 0.1 hours (6.86\%), 0.93 hours (2.11\%), and 2.6 hours (3.59\%) on average for the cases with additional transit times of 0, 0.5, 1, and 2 hours per route respectively compared to the non-DNN method. However, 64\% of the cases in which the non-DNN solution required less time than the DNN-based solution occurred when vehicle capacity was limited to 2 or 4 (Fig. \ref{fig:time_comparison}a-d). When the DNN-based method required less evacuation time than the non-DNN method, it saved 2.1 hours (18.57\%) on average. When the DNN-based method required more evacuation time, it needed 4.51 hours more (13.1\%) on average.

Although deep reinforcement learning can provide more efficient evacuation routing, the time performance boost from implementing the DNN-based solution was not large enough to fully compensate the time penalty from stricter social distancing without other compromises (Figs. \ref{fig:social_distancing_vs_time} and \ref{fig:time_comparison}). Evacuation plans can be modified to accommodate social distancing, but the modifications will require operational changes like increasing the number of vehicles (and decreasing the size of the neighborhood each serves) or extending the evacuation timeline. The magnitude of the required operational changes becomes larger as the social distancing becomes more aggressive. Even if the addition of DNN-based evacuation plans can make a real contribution to efficiency and could, in critical cases, be the difference between a successful evacuation and one that is not completed in the allowed window, it is imperative for disaster managers to re-examine the evacuation timeline and incorporate additional disaster relief resources, such as more emergency vehicles, into any evacuation operation that requires social distancing.

\section*{Discussion}

Artificial intelligence has been employed to solve various hard problems in operations research, computer science, business, healthcare, and other fields. We showed how human-centered AI techniques can augment the efficiency of an evacuation, but its benefit decreases and eventually disappears with stronger requirements for social distancing. The findings from our research are relevant to other disaster evacuations that are based on a registry of people to be collected. The efficiency improvements that come from implementing a DNN-based solution can be substantial, but they may not compensate for the extra time required for adding social distancing to each evacuation route.

In general, the DNN-based solutions were not useful for cases where vehicle capacity was close to the size of a single household (node). In typical CVRP simulations and benchmarking research \cite{kool2018attention, nazari2018reinforcement}, the vehicle capacity is much larger than the demand of each node. In future work, the DNN-based model could be retrained for scenarios with low-capacity vehicles. However, when the vehicle goes to only one house and then returns to the rescue center, all possible solutions collapse to the same performance. Furthermore, with low-capacity vehicles, the total time both DNN-based and non-DNN methods need is far from achieving the threshold, and a modest performance boost from a DNN-based solution would be unlikely to change this pattern, especially in larger neighborhoods (Fig. \ref{fig:social_distancing_vs_time}d).

On the other hand, when vehicle capacity is much larger than the size of each household, multiple households can be combined in a route, and it is non-trivial to come up with an efficient routing strategy. In cases like these, the DNN-based model can reinforce itself by learning from various possible solutions. It is not guaranteed that a DNN-based method can always outperform a non-DNN method, especially in settings without sufficient data (possible routing solutions) for learning and reinforcing the DNN-based model. The limitation of DNN-based method can be a part of the planning for evacuation routing, suggesting cases where DNN-based approaches have and do not have the potential to be helpful.

To accommodate the challenges in disaster evacuations during a pandemic, the approach used in this study could be extended in many ways. For example, both DNN-based and non-DNN methods could be modified to address the split delivery vehicle routing problem (SDVRP -- where an emergency vehicle cannot pick up all people from a house and needs to return for the remaining people). Further, one could add details like intermediate staging points, making this a Vehicle Routing Problem with Pickup and Delivery (VRPPD) in addition to a CVRP. Other variations might allow, for example, elimination of social distancing for family members of fully vaccinated individuals, or a rescue fleet with vehicles of many sizes.

A transit time to a central evacuation center, required in any real-world situation, tends to decrease the advantage of an efficient routing algorithm (Fig. \ref{fig:time_comparison}). For our analysis, it creates an additional burden on the DNN-based approach. This arises from the fact that the DNN algorithm is attempting to minimize the time collecting people, without regard to the number of routes.  But because transit adds a fixed time to each route, it penalizes any solution that requires additional routes. In future work, it may be possible to address this through multi-objective optimization (minimizing both the evacuation time and the number of routes) or to design test neighborhoods with depots located at some distance.

When social distancing is required, evacuations become more complicated, and they require additional resources. Disaster managers will need to accommodate these complexities both through capitalizing on tools for more efficient routing and in other aspects of the evacuations. Specifically, they may want to explore options for (1) extending the evacuation timeline, (2) increasing the number of evacuation vehicles that serves each neighborhood, or (3) partitioning the evacuation area into a larger number of smaller neighborhoods, each served by a single vehicle. In any particular setting, the preferred option will be determined by local resources and constraints. In every location, however, it will be wise to plan ahead for the possibility of an evacuation with social distancing.

One of the durable lessons from COVID-19 is that governments, NGOs, and community members should have all-hazard emergency operations and evacuation plans that consider the interactive effects of both a pandemic and a wide range of possible disasters. Disaster managers should also take advantage of every efficiency they can find, including DNN-based vehicle routing. Consistent with the FEMA suggestion that jurisdictions incorporate social distancing into emergency evacuation during the COVID-19 pandemic \cite{sccovid19}, this study investigates the impacts of social distancing on an evacuation timeline and the number of emergency vehicles required to evacuate a city.

\section*{Methods}

\subsection*{Dataset and problem formulation}

Vehicle Routing Problems (VRPs) in the context of pre-disaster evacuation are both critical and complex \cite{golden2014chapter}. Although we cannot include all real-world constraints, we can roughly formulate the problem of pre-disaster evacuation as the most studied variant of VRPs, the Capacitated Vehicle Routing Problem (CVRP) with a rescue center (typically called \textit{depot}) and a given set of $n$ houses denoted \textit{nodes} or \textit{customers}, $N$ = \{1,2,...,$n$\} \cite{kool2018attention, irnich2014chapter}. 

Traditionally, there are homogeneous vehicles $K$ with the same capacity $C$, where $K$ = $\{1,2,...,\lvert K \rvert \}$, $C$>0. In this study, we calculate total evacuation time based on one emergency vehicle for each neighborhood (with all neighborhoods the same size and being evacuated in parallel), so $K$ is the same vehicle for every route in $K$ in a neighborhood, and $\lvert K \rvert$ is the number of routes required to evacuate each neighborhood. The number of people on the special needs registry to be picked up (typically referred to as \textit{demand}) from each house is $d_j$, where 0 < $d_j$ $\le$ $C$, $j$ $\in$ $N$. Let $E$ = $\{e=\{i,j\}=\{j,i\}:i,j \in \{0,1,2,...,n\}, i \neq j\}$ be the edge set \cite{irnich2014chapter}. An emergency vehicle moving from house $i$ to $j$ incurs travel cost $c_{ij}$ for $\{i,j\} \in E$ or evacuation time in this study \cite{kool2018attention, irnich2014chapter}. We define an integer decision variable $x_{ij}$, where $x_{ij}$=1 if the emergency vehicle moves from house $i$ to $j$ while $x_{ij}$=0 if the emergency vehicle does not move from house $i$ to $j$ \cite{borcinova2017two, irnich2014chapter,kim_2020}.

The vehicle starts from the rescue center, goes to a subset of houses $S$ $\subseteq$ $N$ to pick up people, and finally returns to the rescue center before or until the vehicle capacity $C$ is reached \cite{irnich2014chapter}. This process constitutes one route. The vehicle repeats this process until it collects every resident on the special needs registry, visiting each house only once (Fig. \ref{fig:1} and the other figures in Supplementary information) \cite{irnich2014chapter}. The goal is to evacuate every person on the evacuation registry within the minimum possible evacuation time. Results for a rescue operation consist of the routes utilized by this vehicle, evacuation time, the number of routes, and other detailed information about each route (see Supplementary information). The formulation of the CVRP can be expressed mathematically as follows \cite{borcinova2017two, irnich2014chapter,kim_2020}.

\begin{equation} \label{eq:3}{Minimize\ \sum_{k\in K} \sum_{i=0}^n \sum_{j=0, i\neq j}^n c_{ij} x_{ij}^k, } 
\end{equation}

\begin{equation} \label{eq:4}{\sum_{k\in K} \sum_{i=0}^n \sum_{j=1, i\neq j}^n x_{ij}^k = 1,}
\end{equation}

\begin{equation} \label{eq:5}{\sum_{j\in N} x_{0j}^k = 1, \ \ \ \forall k \in K,}
\end{equation}

\begin{equation} \label{eq:6}{\sum_{i=0,i \neq j}^n x_{ij}^k = \sum_{i=0}^n x_{ji}^k, \ \ \ j=\{0,1,...,n\}, \forall k \in K,}
\end{equation}

\begin{equation} \label{eq:7}{\sum_{i=0}^n \sum_{j=1,i \neq j}^n d_j x_{ij}^k \leq C, \ \ \ \forall k \in K,}
\end{equation}

\begin{equation} \label{eq:8}{\sum_{k\in K} \sum_{i\in S} \sum_{j\in S,i \neq j} x_{ij}^k \leq \lvert S \rvert - 1, \ \ \ \forall S \subseteq \{1,2,...,n\},}
\end{equation}

\begin{equation} \label{eq:9}{x_{ij}^k \in \{0,1\}, \ \ \ \forall k \in K, \{i,j\} \in E}
\end{equation}

According to the parameters we defined earlier \cite{borcinova2017two, irnich2014chapter,kim_2020}, (\ref{eq:3}) is the objective function which minimizes the overall costs (total evacuation time) for the neighborhood. Constraint (\ref{eq:4}) makes sure that each house is visited only once. Constraint (\ref{eq:5}) indicates that the the emergency vehicle can leave the rescue center only once for each route in $K$. Constraint (\ref{eq:6}) requires that the number of the emergency vehicles arriving at and leaving each house or the rescue center is the same (1 in this study). Constraint (\ref{eq:7}) ensures that the number of people picked by the emergency vehicle in a single route is no more than the vehicle capacity $C$. (\ref{eq:8}) is called the subtour elimination constraint, which avoids any route that is disconnected from the rescue center\cite{borcinova2017two,irnich2014chapter,semet2014chapter}. Lastly, constraint (\ref{eq:9}) denotes that the integer decision variable should be either 1 (visited) or 0 (not visited).

To analyze the efficacy of pre-disaster evacuations using Deep Reinforcement Learning (DNN-based solution) and Sweep Algorithm (Non-DNN solution), we consider four datasets in this study.

\subsubsection*{Dataset selection}
    In our experiments, we used the A-n36-k5, A-n53-k7, A-n69-k9 datasets, commonly used in studies of capacitated vehicle routing problems \cite{cvrplib, augerat1995computational}. Each dataset contains a unique set of coordinates of one depot and several nodes in a two-dimensional Cartesian coordinate system, plus the demand of each node and vehicle capacity. Since we are interested in exploring how size of neighborhood and social distancing policy could impact the total time and the number of routes in evacuation operations, we selected the first 20 locations in the A-n36-k5 dataset as our Dataset 1. We used all 35 locations in A-n36-k5 as our Dataset 2. Our Dataset 3 and 4 are  A-n53-k7, A-n69-k9 datasets which contain 52 and 68 houses respectively. To accommodate the possibility that the depot is outside the neighborhood, we allowed for a series of fixed times (0, 0.5, 1.0, and 2.0 hours) for transit to and from a central depot.

\subsubsection*{Dataset formation}
    In structuring the problem, we use the city of New Orleans, Louisiana, USA, as the reference case. New Orleans experienced massive losses and disruption in Hurricane Katrina (2005) \cite{dyson2010come, national2009new, richardson2008natural, kettl2006risk, banipal2006strategic} and has implemented comprehensive plans to prepare for a future major hurricane. While our analytical framework is general, details like the average number of people in a household and the length of the time window for completing the evacuation are specific to New Orleans.
    
    In New Orleans, the elderly, disabled, and those at high risk for severe illness from COVID-19 can sign up online or call 311 for the Special Needs Registry (SNR) before a mandatory evacuation order is announced \cite{plan, schrilla2019evacuation}. The 72-hour evacuation timeline announced by the City of New Orleans specifies a window of 42 hours to collect evacuees from their houses and transport them to the Smoothie King Center, which serves as a transfer and processing center, and from there, evacuees will board a bus to state or federal shelters in other cities \cite{plan}. Finally, the city government will bring evacuees back to their homes or local shelters once it becomes safe to return to New Orleans \cite{plan}. As of February 23, 2021, the special needs registry for New Orleans included approximately 4,000 individuals. \cite{schrilla_2021}.

    We considered four datasets, each consisting of a depot (rescue center) and several nodes (houses) randomly distributed within a flat 2D grid world. The distance metric is Manhattan distance, as in an urban area laid out in blocks (see Supplementary Fig. \ref{fig:cvrp_manhattan} online). Each dataset is in the shape of a square (about 3 km x 3 km) \cite{schrilla_2021_april}. The land area of New Orleans is \SI{438.8}{\kilo\metre\squared} (\SI{169.42}{mile^2}) \cite{census} so the size of each dataset represents about 2\%\ of the area of the city. 
    
    To enhance the efficiency of computation, we normalize the neighborhood as a 1 x 1 box, and transform the normalized distance back to actual distance to get total evacuation time.

To estimate the number of people in each house, we used the average household size (2014-2018), which is 2.44 persons per household in New Orleans, based on the American Community Survey (ACS) of U.S. Census Bureau \cite{census}, assuming a normal distribution with mean=2.44 and standard deviation=0.5. If the household size is not an integer, it is rounded to the nearest integer. So, the household size ranges from 1-4 people. For simplicity, we assume that, if one person in a house is on the special needs registry, then all of the people in that house are on the registry as well. Supplementary Table \ref{table:dataset} contains the summary of the datasets used in our study. 

We ran the sweep algorithm and the pre-trained deep reinforcement learning model on a GPU NVIDIA Tesla K80, which enabled us to complete all experiments within one second with a batch size of 256 \cite{kool2018attention}. For our conceptual model of a pre-planned evacuation map, calculation time is not relevant, but the possibility of quick calculation is consistent with real-time adjustments. In particular, a timely and scalable solution is of vital importance in real-world disaster response and evacuation route planning.

\subsection*{Emergency vehicle and social distancing}
We chose the average speed of 8 km/h (equivalent to 5 mph) for our simulation, based on the idea that emergency vehicles usually move slowly during evacuation operations, especially when the elderly and people with disabilities need extra time to get on and off the vehicles. To investigate how social distancing in an emergency vehicle impacts the total time and the number of routes in disaster evacuation operation, we consider a nominal evacuation vehicle as a bus that seats up to 64 people. Social distancing could decrease that to 32, 16, 8, 4, or 2 people per emergency vehicle. The analysis can also be applied to a nominal capacity of 32, 16, 8, or 4, with an increasingly restricted range of social distancing. To our knowledge, this study is the first research which incorporates FEMA's official guidelines \cite{sccovid19} into the investigation of the impacts of social distancing on emergency evacuation. In light of FEMA's guidance on Recommended Evacuee Queuing and Boarding Process \cite{sccovid19}, with social distancing protocols, only up to 25-28 passengers are allowed in a 56-passenger motor coach, which is the most widely available motor coach. In other words, the vehicle capacity decreases to half of its original value when social distancing is applied. All of the specifics like the size of the neighborhoods, vehicle speed, and vehicle capacity are reasonable, chosen to illustrate the general issues.

\subsection*{Evacuees and evacuation planning}
This is a model for a pre-planned evacuation, with a map of locations and the number of people per location known in advance. Such a map could be based on a registry of advance requests for evacuation assistance, or it could be developed in parallel with the forecast leading to an evacuation. With this concept, we know in advance the spatial distribution of the demand in the capacitated vehicle routing problem. In addition, we considered the rescue center to have unlimited accommodation capacity. Consistent with the Guide of City-Assisted Evacuation (CAE) for Hurricanes in New Orleans, people transported from their houses to the rescue center will be moved from there to be treated by appropriate emergency health care services at other disaster relief centers, such as state or federal long-term shelters \cite{evacuspot, plan}.

\subsection*{Algorithm design}
In this section, we described the general design of the DNN-based and non-DNN algorithms in this study.

\subsubsection*{Non-DNN solution--Sweep Algorithm}
The sweep algorithm is a computationally efficient non-DNN solution typically used in real-world evacuations, business logistics, and supply chain management \cite{swamy2017hurricane, ballou2007business}. It starts with an arbitrary line from the depot (the rescue center). The order of houses to be visited by an emergency vehicle is determined by sweeping this line counter-clock wise and adding houses one by one when the line intersects these houses. In addition, the emergency vehicle must return to the rescue center when it reaches its passenger limit. After sweeping the line for 360 degrees, the evacuation operation is complete.

\subsubsection*{DNN-based solution--Deep Reinforcement Learning}
In this study, we selected deep reinforcement learning as the DNN-based solution. Deep reinforcement learning is good at searching for optimal solutions in a relatively short period of time and is well-known for its capacity of adaptively resolving similarly complex problems, such as the Game of Go, robotics, and computer games. In particular, we used the Attention Model (AM) \cite{kool2018attention} because this algorithm outperformed several common baseline algorithms and models for various routing problems, including the Capacitated Vehicle Routing Problem (CVRP). The Attention Model integrates the REINFORCE algorithms \cite{williams1992simple} with greedy rollout baseline to the attention-based Transformer model \cite{vaswani2017attention} and the variant of Graph Attention Networks (GATs) \cite{velickovic2018graph} whose Convolutional Neural Networks (CNNs) with masked self-attention layers analyze graph-structured data efficiently. For all of our experiments of deep reinforcement learning, we adapted the algorithm designs in \cite{kool2018attention} to our scenarios of evacuation with social distancing.

\subsection*{Evaluation of DNN-based and non-DNN solutions}

According to the 72-hour hurricane evacuation timeline announced by New Orleans \cite{plan}, there will be only 42 hours to pick up evacuees (Fig. \ref{fig:timeline}). To evaluate the performance of DNN-based and non-DNN solutions, we used 24 and 42 hours as the thresholds of categorizing the results into three levels of performance: satisfactory (< 24 hours), borderline (24-42 hours), and not allowed (> 42 hours) (Fig. \ref{fig:timeline}).

\section*{Data availability}
The datasets generated during and/or analyzed during the current study are available in the Capacitated Vehicle Routing Problem Library
(\url{http://vrp.galgos.inf.puc-rio.br/index.php/en/}) and from the corresponding author on reasonable request.

\printbibliography

\section*{Acknowledgements}
\textit{Funding:} This research was supported by Stanford Woods Institute for the Environment, Department of Civil \& Environmental Engineering at Stanford University, Department of Earth System Science at Stanford University, Microsoft AI for Earth Program, and Stanford RISE (Respond. Innovate. Scale. Empower.) COVID-19 Crisis Response Research Grant and Fellowship.\\
\textit{Insights:} We appreciate those who provide their first-hand insights and work experiences in disaster response and evacuation planning:\\
1. Thomas Schrilla--Planner, Office of Homeland Security and Emergency Preparedness, City of New Orleans.\\
2. Ann Herosy--Team Lead of the Emergency Management Team and the Community Partners Team, American Red Cross of the Silicon Valley.\\
3. Luke Beckman--Division Disaster State Relations Director for California, American Red Cross Pacific Division.\\
4. Keith Perry--University Emergency Manager, Assistant Director, Department of Environmental Health and Safety (EH\&S), Stanford University. \\
\textit{Technical Supports:} We thank the City of New Orleans, the Office of Homeland Security and Emergency Preparedness (NOHSEP) for permitting us to use its hurricane evacuation timeline and publish as shown in Fig. \ref{fig:timeline}, and cite the real-world evacuation settings, such as the number of evacuees (4,000 individuals) and spatial dimension of one neighborhood (about 3 km x 3 km), in their latest evacuation planning for the 2021 hurricane season. In addition, we thank Abhijeet Phatak and Laura Domine for providing advice and exploratory assistance in the preliminary design of this study. We thank Professor Sarah Fletcher and Laura Domine for rigorously reviewing our manuscript, and providing feedback and suggestions in detail. We also thank Professor Emma Brunskill and Eley Ng for their guidance on some technical issues of deep reinforcement learning. We appreciate that Wouter Kool helped us figure out technical bugs and clarify some issues of DNN-based models when we adapted the ideas in \cite{kool2018attention} to our scenarios of flood evacuation with social distancing. Last but not least, we thank Dr. Rishi Mediratta, Jack Scala, and Nathaniel Braun on the Stanford teaching team of PEDS 220: COVID-19 Elective for their support and guidance on the reciprocal community-based research partnership between the City of New Orleans and Stanford University.

\section*{Author contributions}
Y.L.T., P.K.K., and C.B.F. designed the research; all authors developed and refined the methodologies of analyses; Y.L.T. and C.R. performed the analyses; Y.L.T., P.K.K., and C.B.F. interpreted results. Y.L.T. wrote the first draft of the manuscript. Y.L.T., P.K.K., and C.B.F. revised the manuscript. All authors reviewed the final version of the manuscript.

\section*{Ethics declarations}
\subsection*{Competing interests}
The authors declare no competing interests.

\section*{Disclaimer}
This disclaimer informs readers that this study should only serve as their own references rather than official guidelines for any jurisdictions. The scenarios of simulations performed in this study are only examples for the purpose of academic research. Any actions, including but not limited to decisions, policies, and studies, taken based on any part of this study is the sole liability of readers, not authors in this study.

\section*{Supplementary information}
\subsection*{Routing experiments}
\textit 
This section included all of the routing figures generated from our DNN-based and non-DNN solutions. In each figure, a legend box shows the detailed information about that experiment. For example, in Supplementary Fig. \ref{s1}, the legend box of the DNN-based Solution is in the format of “R0, \#14, c 37/64, t: 1.79 hours”. R0 is the first route, R1 is the second, etc. “\#14” means that an emergency vehicle visits 14 houses in that route. “c 37/64” indicates that the emergency vehicle whose passenger capacity is 64 people picks up 37 people in route R0. Finally, “t: 1.79 hours” means that the emergency vehicle spends 1.79 hours in route R0.
To make the routing pattern easy to read, we hide the first and the last line going from and back to the black square, which represents the transition point for transit to the depot. For each experiment, we showed the total time and the number of routes at the top of each figure. All of the figures in this supplementary information section are for vehicle capacity = 64, 32, 16, 8, 4, or 2 people per emergency vehicle and transit time = 0.
In our study, we used the typical CVRP setting, meaning the emergency vehicle would only visit each house once. For cases where vehicle capacity was 2, to deal with some houses having more than two people, we split the simulations into two parts. In Part I, the vehicle can take up to two people from each household, as shown in Supplementary Fig. \ref{s6}. People left in each house were picked up in Part II, as shown in Supplementary Fig. \ref{s7}.

\begin{table}[H]
\caption{\textbf{Summary of datasets\cite{cvrplib, augerat1995computational}.}\label{table:dataset}.} 
\includegraphics[width=0.5\textwidth]{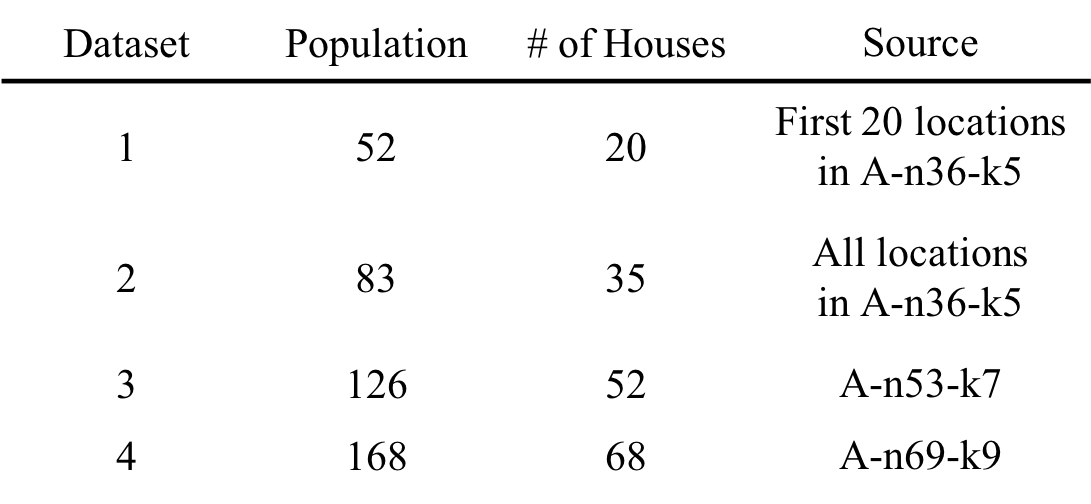}\centering
\centering

\end{table}

\setcounter{figure}{0}   

\subsection*{Dataset 1: 52 people and 20 houses}
\begin{figure}[H]
	\centering
	\begin{subfigure}{0.48\textwidth} %
		\includegraphics[width=\textwidth]{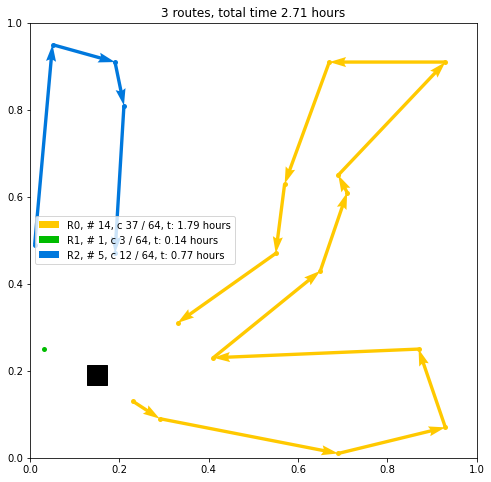}
		\caption{DNN-based Solution} %
	\end{subfigure}
	\vspace{0.1cm} %
	\begin{subfigure}{0.48\textwidth} %
		\includegraphics[width=\textwidth]{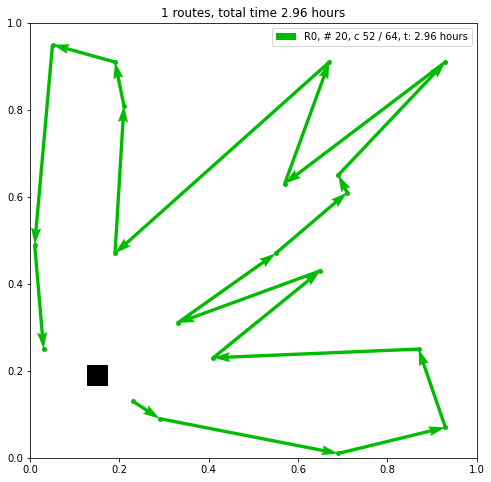}
		\caption{Non-DNN Solution} %
	\end{subfigure}
	\caption{\textbf{Social distancing policy: passenger capacity=64 people.}}\label{s1} %
\end{figure}

\begin{figure}[H]
	\centering
	\begin{subfigure}{0.48\textwidth} %
		\includegraphics[width=\textwidth]{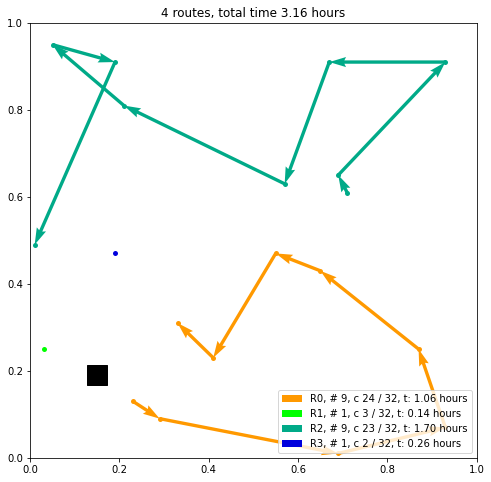}
		\caption{DNN-based Solution} %
	\end{subfigure}
	\vspace{0.1cm} %
	\begin{subfigure}{0.48\textwidth} %
		\includegraphics[width=\textwidth]{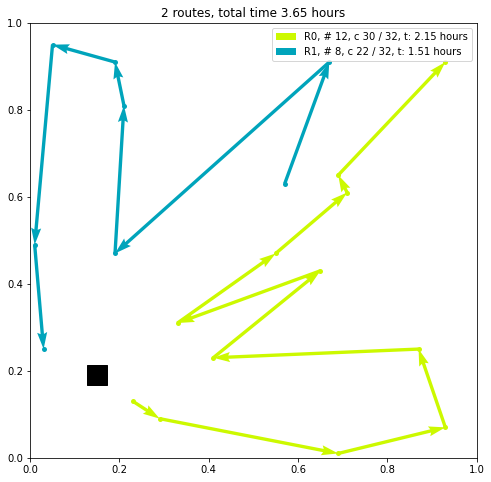}
		\caption{Non-DNN Solution} %
	\end{subfigure}
	\caption{\textbf{Social distancing policy: passenger capacity=32 people.}}\label{s2} %
\end{figure}

\begin{figure}[H]
	\centering
	\begin{subfigure}{0.48\textwidth} %
		\includegraphics[width=\textwidth]{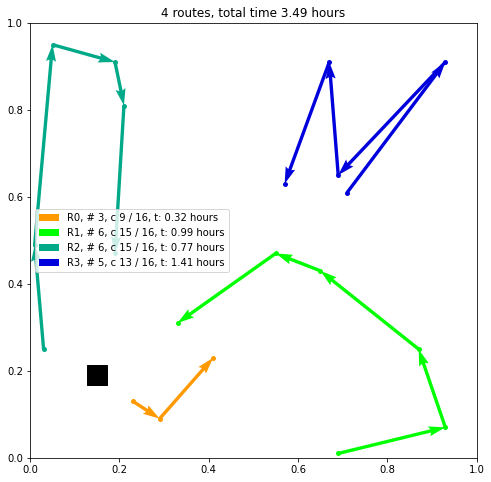}
		\caption{DNN-based Solution} %
	\end{subfigure}
	\vspace{0.1cm} %
	\begin{subfigure}{0.48\textwidth} %
		\includegraphics[width=\textwidth]{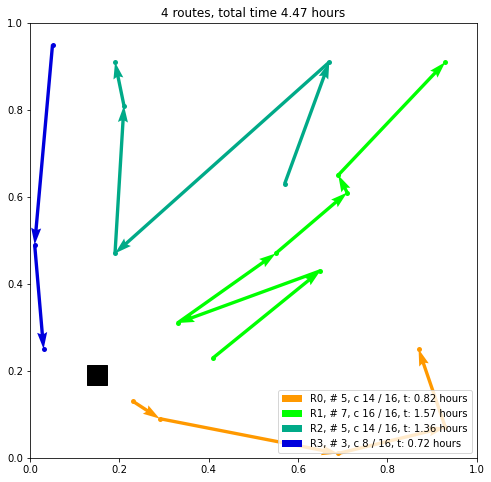}
		\caption{Non-DNN Solution} %
	\end{subfigure}
	\caption{\textbf{Social distancing policy: passenger capacity=16 people.}}\label{s3} %
\end{figure}

\begin{figure}[H]
	\centering
	\begin{subfigure}{0.48\textwidth} %
		\includegraphics[width=\textwidth]{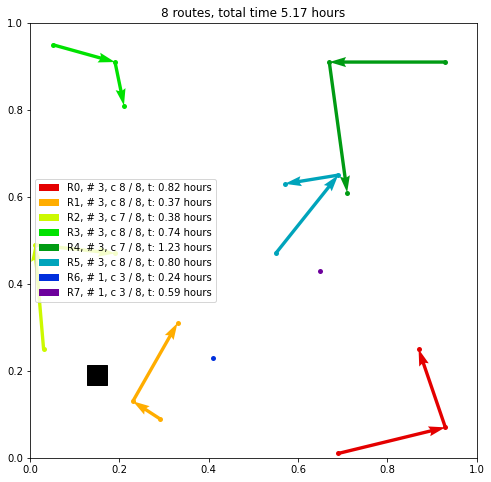}
		\caption{DNN-based Solution} %
	\end{subfigure}
	\vspace{0.1cm} %
	\begin{subfigure}{0.48\textwidth} %
		\includegraphics[width=\textwidth]{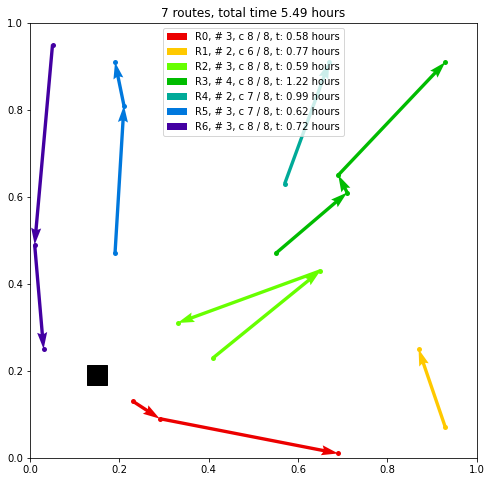}
		\caption{Non-DNN Solution} %
	\end{subfigure}
	\caption{\textbf{Social distancing policy: passenger capacity=8 people.}}\label{s4} %
\end{figure}

\begin{figure}[H]
	\centering
	\begin{subfigure}{0.48\textwidth} %
		\includegraphics[width=\textwidth]{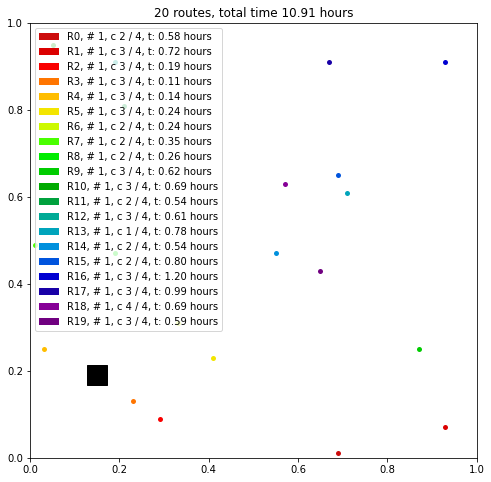}
		\caption{DNN-based Solution} %
	\end{subfigure}
	\vspace{0.1cm} %
	\begin{subfigure}{0.48\textwidth} %
		\includegraphics[width=\textwidth]{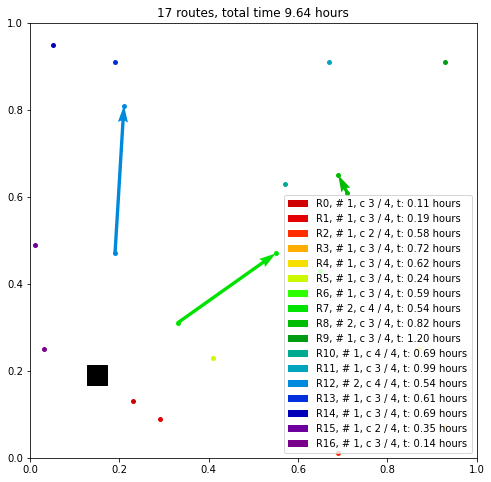}
		\caption{Non-DNN Solution} %
	\end{subfigure}
	\caption{\textbf{Social distancing policy: passenger capacity=4 people.}}\label{s5} %
\end{figure}

\begin{figure}[H]
	\centering
	\begin{subfigure}{0.48\textwidth} %
		\includegraphics[width=\textwidth]{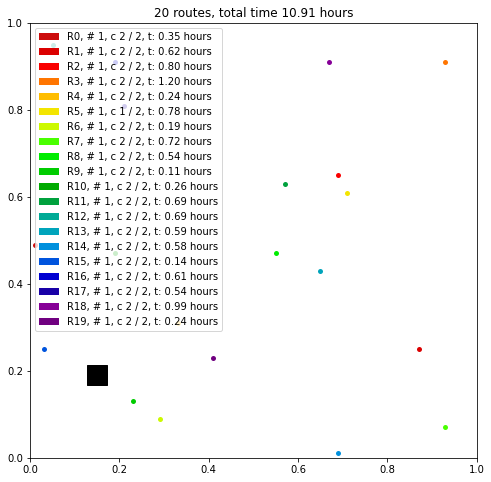}
		\caption{DNN-based Solution} %
	\end{subfigure}
	\vspace{0.1cm} %
	\begin{subfigure}{0.48\textwidth} %
		\includegraphics[width=\textwidth]{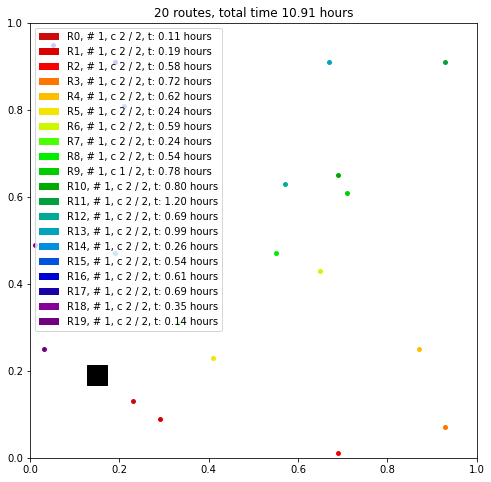}
		\caption{Non-DNN Solution} %
	\end{subfigure}
	\caption{\textbf{Social distancing policy: passenger capacity=2 people, Part I.}}\label{s6} %
\end{figure}

\begin{figure}[H]
	\centering
	\begin{subfigure}{0.48\textwidth} %
		\includegraphics[width=\textwidth]{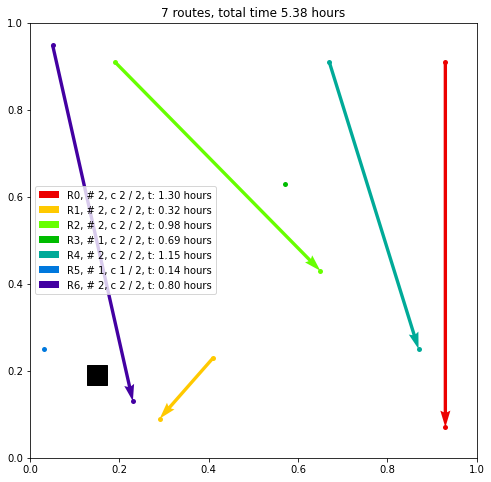}
		\caption{DNN-based Solution} %
	\end{subfigure}
	\vspace{0.1cm} %
	\begin{subfigure}{0.48\textwidth} %
		\includegraphics[width=\textwidth]{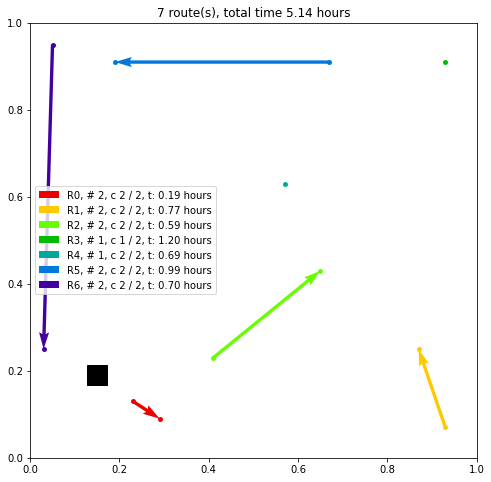}
		\caption{Non-DNN Solution} %
	\end{subfigure}
	\caption{\textbf{Social distancing policy: passenger capacity=2 people, Part II.}}\label{s7} %
\end{figure}

\subsection*{Dataset 2: 83 people and 35 houses}
\begin{figure}[H]
	\centering
	\begin{subfigure}{0.48\textwidth} %
		\includegraphics[width=\textwidth]{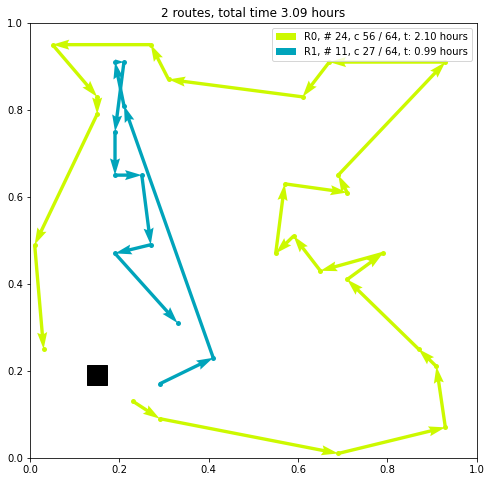}
		\caption{DNN-based Solution} %
	\end{subfigure}
	\vspace{0.1cm} %
	\begin{subfigure}{0.48\textwidth} %
		\includegraphics[width=\textwidth]{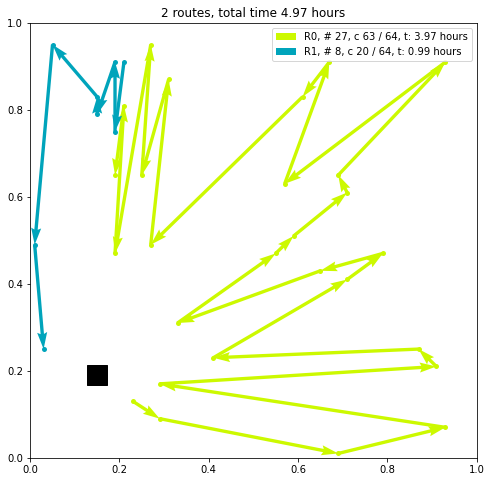}
		\caption{Non-DNN Solution} %
	\end{subfigure}
	\caption{\textbf{Social distancing policy: passenger capacity=64 people.}}\label{s8} %
\end{figure}

\begin{figure}[H]
	\centering
	\begin{subfigure}{0.48\textwidth} %
		\includegraphics[width=\textwidth]{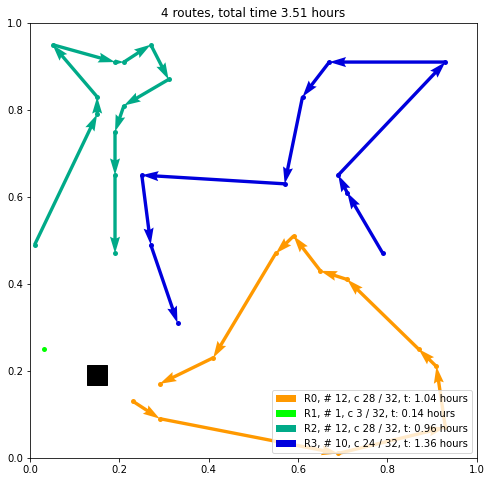}
		\caption{DNN-based Solution} %
	\end{subfigure}
	\vspace{0.1cm} %
	\begin{subfigure}{0.48\textwidth} %
		\includegraphics[width=\textwidth]{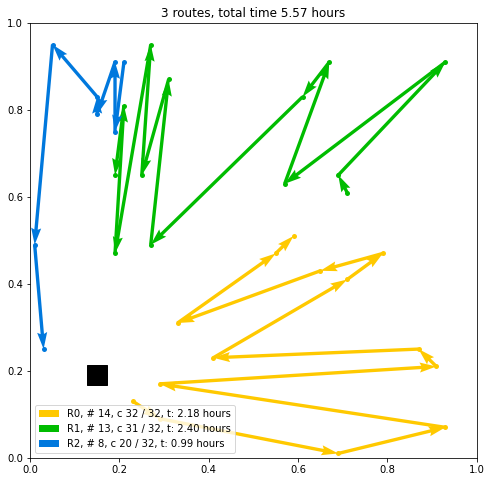}
		\caption{Non-DNN Solution} %
	\end{subfigure}
	\caption{\textbf{Social distancing policy: passenger capacity=32 people.}}\label{s9} %
\end{figure}

\begin{figure}[H]
	\centering
	\begin{subfigure}{0.48\textwidth} %
		\includegraphics[width=\textwidth]{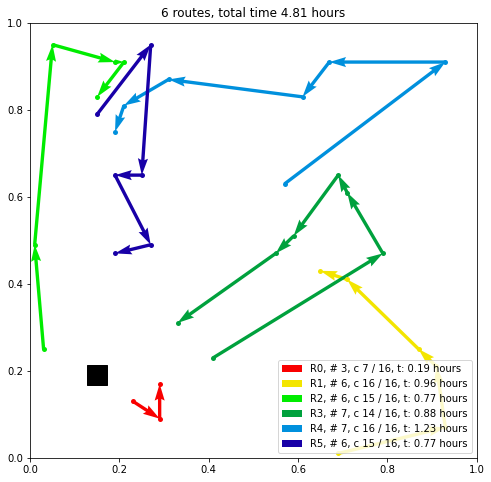}
		\caption{DNN-based Solution} %
	\end{subfigure}
	\vspace{0.1cm} %
	\begin{subfigure}{0.48\textwidth} %
		\includegraphics[width=\textwidth]{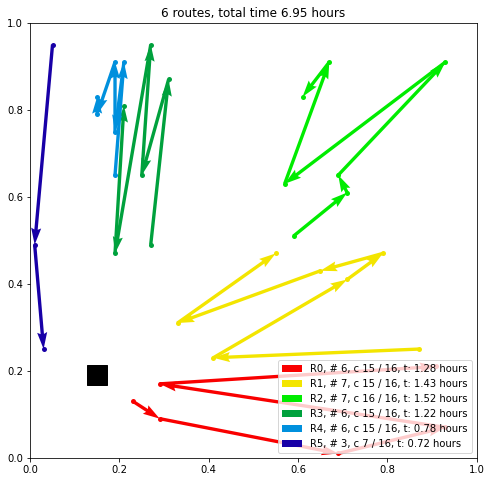}
		\caption{Non-DNN Solution} %
	\end{subfigure}
	\caption{\textbf{Social distancing policy: passenger capacity=16 people.}}\label{s10} %
\end{figure}

\begin{figure}[H]
	\centering
	\begin{subfigure}{0.48\textwidth} %
		\includegraphics[width=\textwidth]{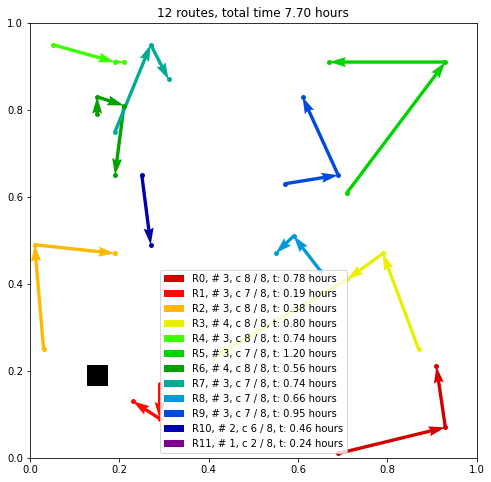}
		\caption{DNN-based Solution} %
	\end{subfigure}
	\vspace{0.1cm} %
	\begin{subfigure}{0.48\textwidth} %
		\includegraphics[width=\textwidth]{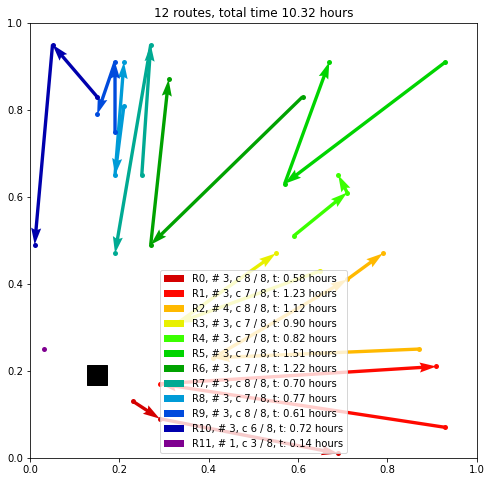}
		\caption{Non-DNN Solution} %
	\end{subfigure}
	\caption{\textbf{Social distancing policy: passenger capacity=8 people.}}\label{s11} %
\end{figure}

\begin{figure}[H]
	\centering
	\begin{subfigure}{0.48\textwidth} %
		\includegraphics[width=\textwidth]{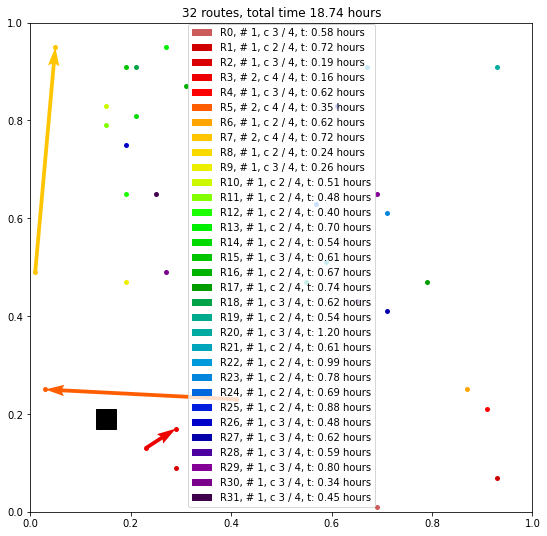}
		\caption{DNN-based Solution} %
	\end{subfigure}
	\vspace{0.1cm} %
	\begin{subfigure}{0.48\textwidth} %
		\includegraphics[width=\textwidth]{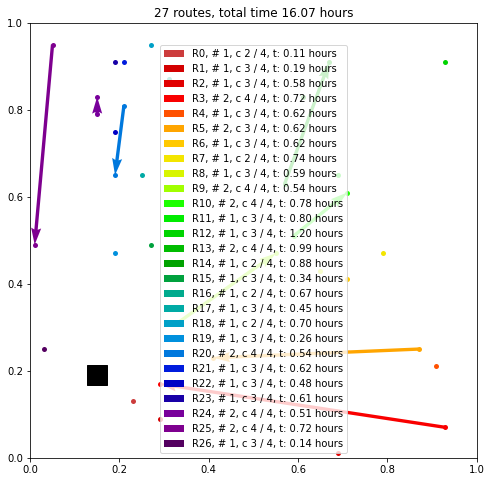}
		\caption{Non-DNN Solution} %
	\end{subfigure}
	\caption{\textbf{Social distancing policy: passenger capacity=4 people.}}\label{s12} %
\end{figure}

\begin{figure}[H]
	\centering
	\begin{subfigure}{0.48\textwidth} %
		\includegraphics[width=\textwidth]{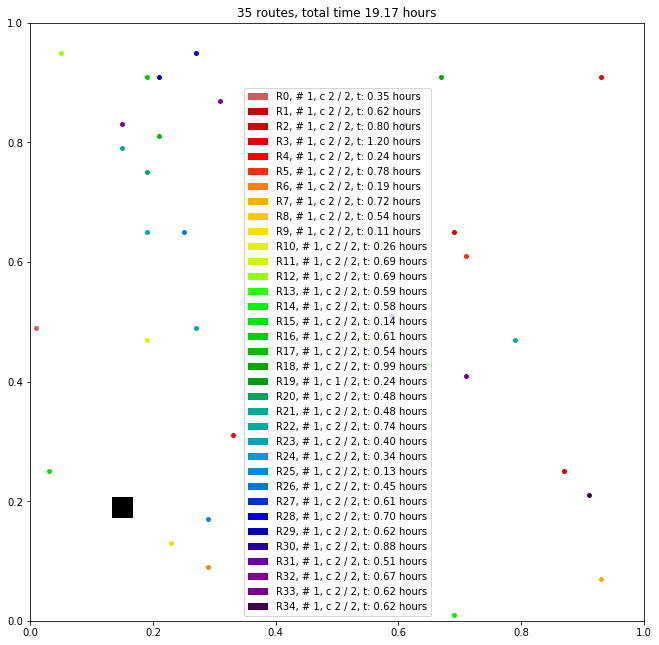}
		\caption{DNN-based Solution} %
	\end{subfigure}
	\vspace{0.1cm} %
	\begin{subfigure}{0.48\textwidth} %
		\includegraphics[width=\textwidth]{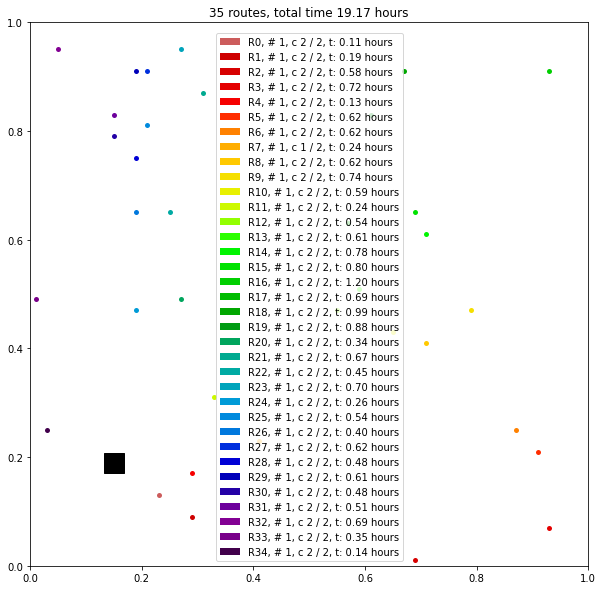}
		\caption{Non-DNN Solution} %
	\end{subfigure}
	\caption{\textbf{Social distancing policy: passenger capacity=2 people, Part I.}}\label{s13} %
\end{figure}

\begin{figure}[H]
	\centering
	\begin{subfigure}{0.48\textwidth} %
		\includegraphics[width=\textwidth]{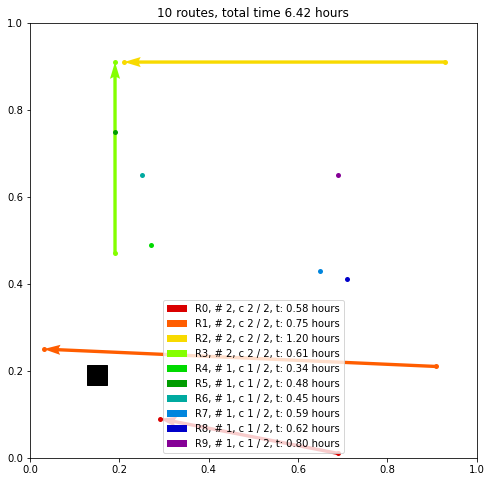}
		\caption{DNN-based Solution} %
	\end{subfigure}
	\vspace{0.1cm} %
	\begin{subfigure}{0.48\textwidth} %
		\includegraphics[width=\textwidth]{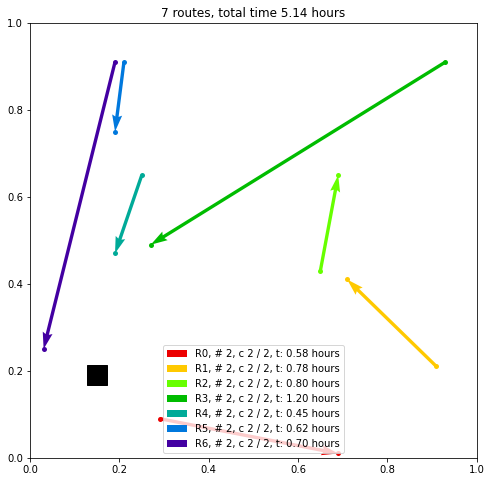}
		\caption{Non-DNN Solution} %
	\end{subfigure}
	\caption{\textbf{Social distancing policy: passenger capacity=2 people, Part II.}}\label{s14} %
\end{figure}

\subsection*{Dataset 3: 126 people and 52 houses}
\begin{figure}[H]
	\centering
	\begin{subfigure}{0.48\textwidth} %
		\includegraphics[width=\textwidth]{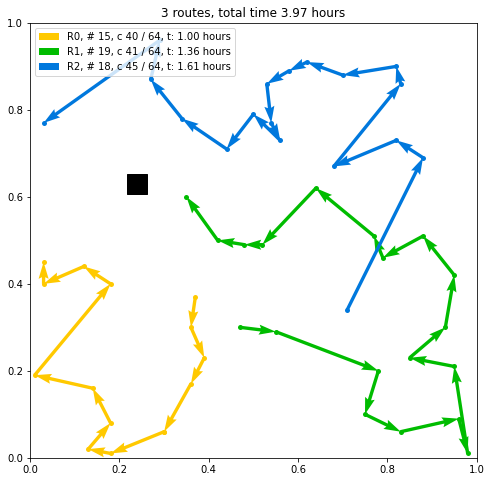}
		\caption{DNN-based Solution} %
	\end{subfigure}
	\vspace{0.1cm} %
	\begin{subfigure}{0.48\textwidth} %
		\includegraphics[width=\textwidth]{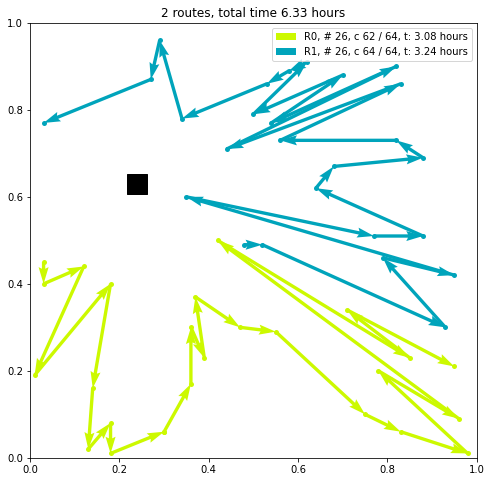}
		\caption{Non-DNN Solution} %
	\end{subfigure}
	\caption{\textbf{Social distancing policy: passenger capacity=64 people.}}\label{s15} %
\end{figure}

\begin{figure}[H]
	\centering
	\begin{subfigure}{0.48\textwidth} %
		\includegraphics[width=\textwidth]{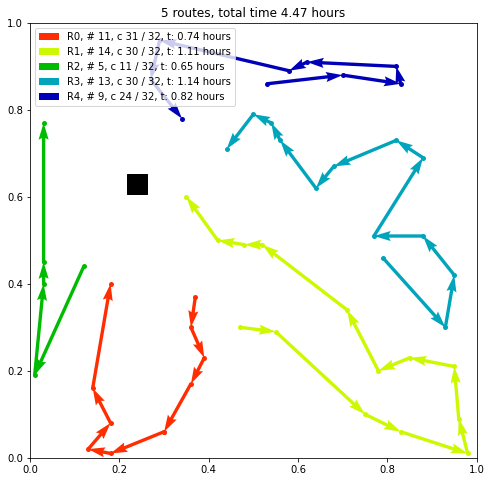}
		\caption{DNN-based Solution} %
	\end{subfigure}
	\vspace{0.1cm} %
	\begin{subfigure}{0.48\textwidth} %
		\includegraphics[width=\textwidth]{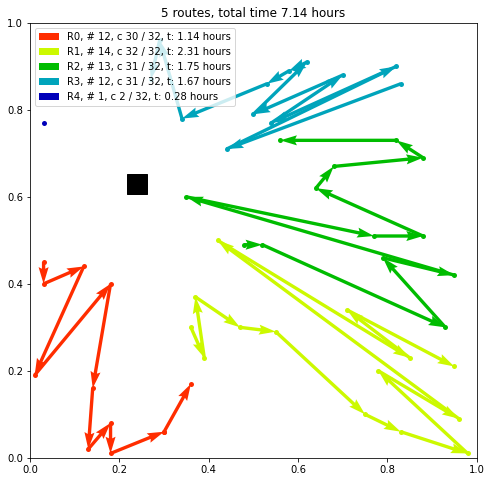}
		\caption{Non-DNN Solution} %
	\end{subfigure}
	\caption{\textbf{Social distancing policy: passenger capacity=32 people.}}\label{s16} %
\end{figure}

\begin{figure}[H]
	\centering
	\begin{subfigure}{0.48\textwidth} %
		\includegraphics[width=\textwidth]{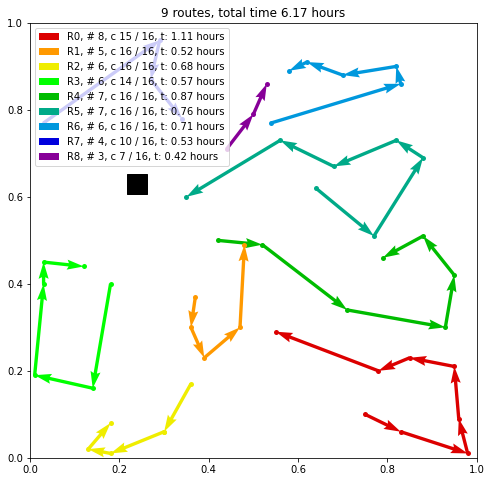}
		\caption{DNN-based Solution} %
	\end{subfigure}
	\vspace{0.1cm} %
	\begin{subfigure}{0.48\textwidth} %
		\includegraphics[width=\textwidth]{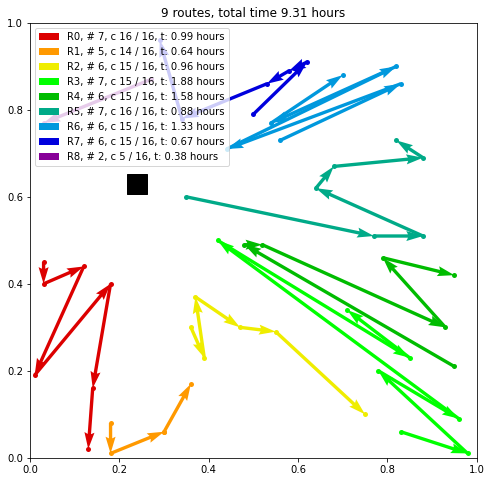}
		\caption{Non-DNN Solution} %
	\end{subfigure}
	\caption{\textbf{Social distancing policy: passenger capacity=16 people.}}\label{s17} %
\end{figure}

\begin{figure}[H]
	\centering
	\begin{subfigure}{0.48\textwidth} %
		\includegraphics[width=\textwidth]{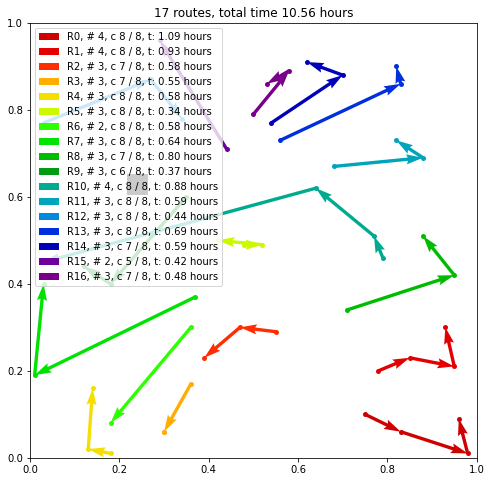}
		\caption{DNN-based Solution} %
	\end{subfigure}
	\vspace{0.1cm} %
	\begin{subfigure}{0.48\textwidth} %
		\includegraphics[width=\textwidth]{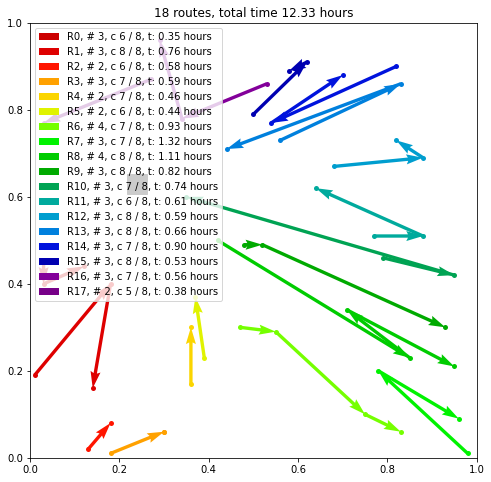}
		\caption{Non-DNN Solution} %
	\end{subfigure}
	\caption{\textbf{Social distancing policy: passenger capacity=8 people.} }\label{s18} %
\end{figure}

\begin{figure}[H]
	\centering
	\begin{subfigure}{0.48\textwidth} %
		\includegraphics[width=\textwidth]{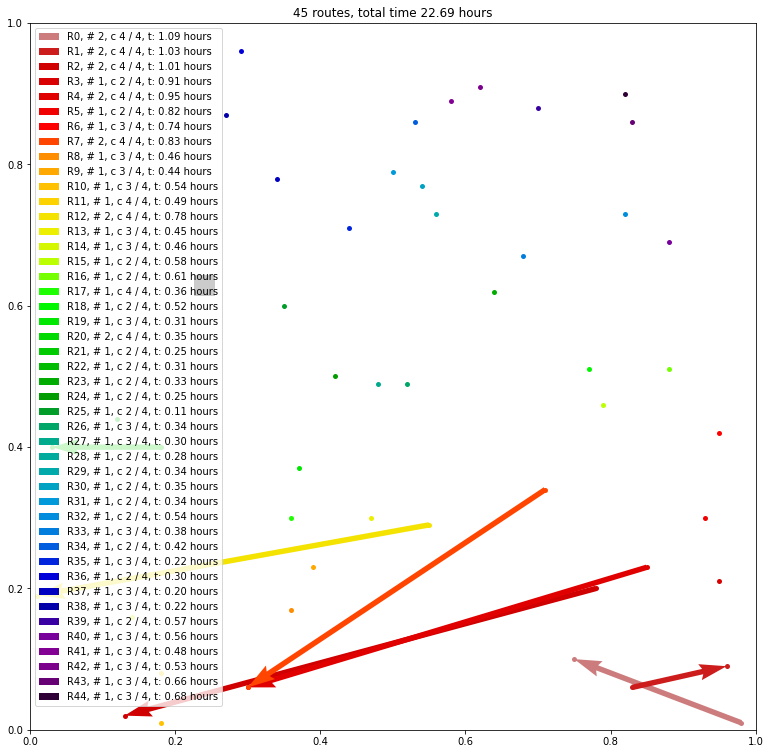}
		\caption{DNN-based Solution} %
	\end{subfigure}
	\vspace{0.1cm} %
	\begin{subfigure}{0.48\textwidth} %
		\includegraphics[width=\textwidth]{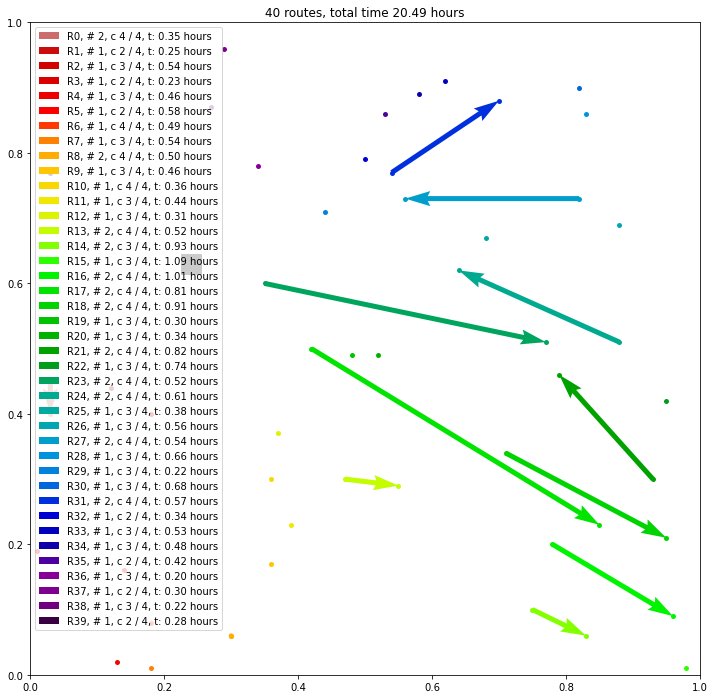}
		\caption{Non-DNN Solution} %
	\end{subfigure}
	\caption{\textbf{Social distancing policy: passenger capacity=4 people.} }\label{s19} %
\end{figure}

\begin{figure}[H]
	\centering
	\begin{subfigure}{0.48\textwidth} %
		\includegraphics[width=\textwidth]{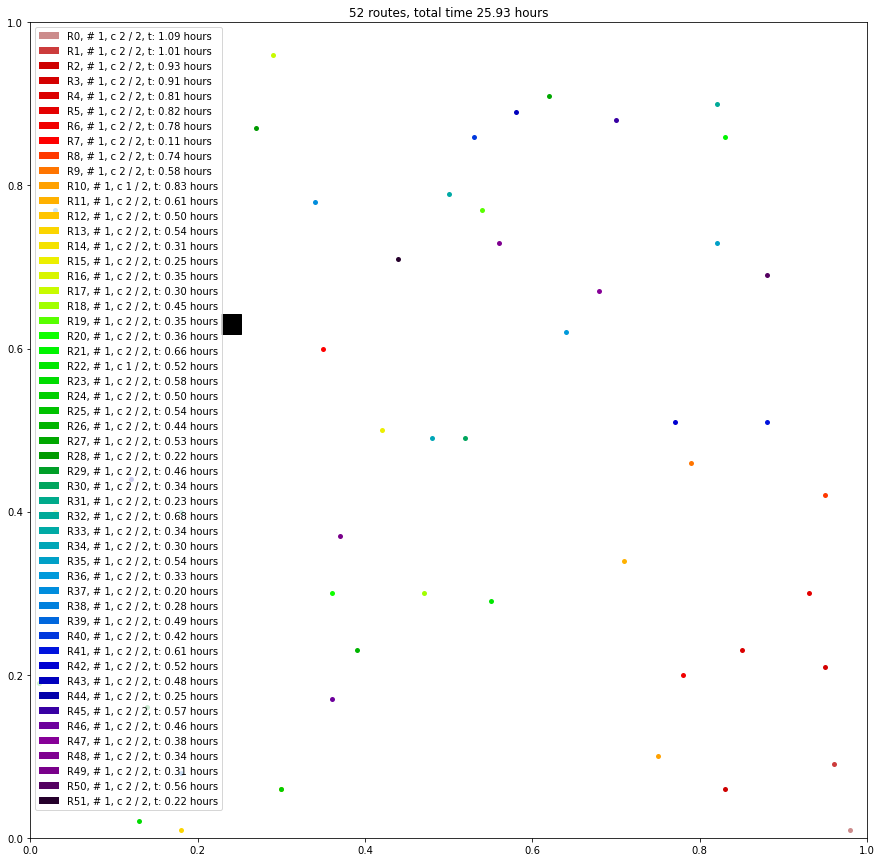}
		\caption{DNN-based Solution} %
	\end{subfigure}
	\vspace{0.1cm} %
	\begin{subfigure}{0.48\textwidth} %
		\includegraphics[width=\textwidth]{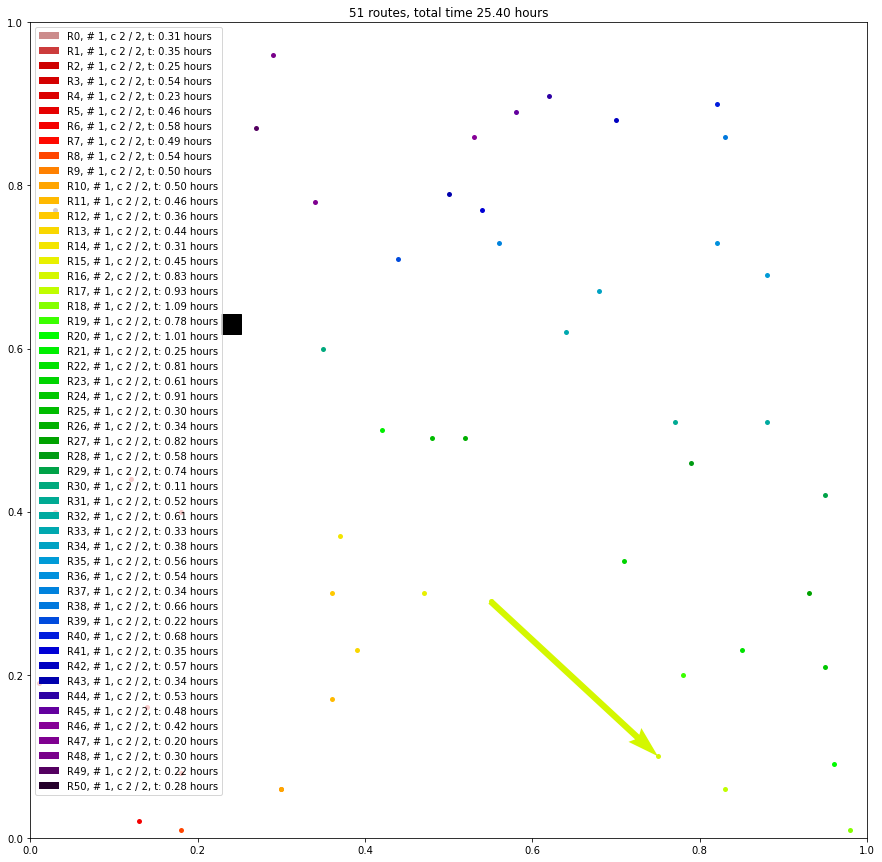}
		\caption{Non-DNN Solution} %
	\end{subfigure}
	\caption{\textbf{Social distancing policy: passenger capacity=2 people, Part I.}}\label{s20} %
\end{figure}

\begin{figure}[H]
	\centering
	\begin{subfigure}{0.48\textwidth} %
		\includegraphics[width=\textwidth]{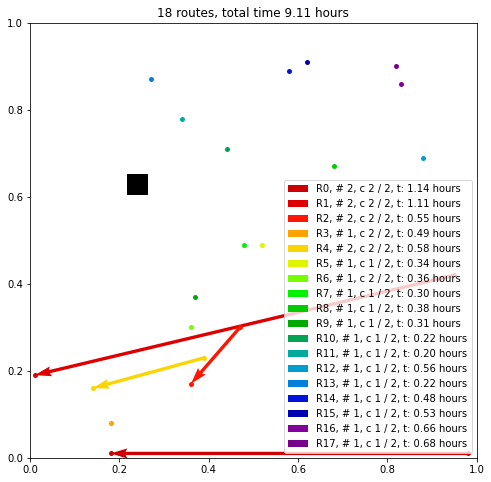}
		\caption{DNN-based Solution} %
	\end{subfigure}
	\vspace{0.1cm} %
	\begin{subfigure}{0.48\textwidth} %
		\includegraphics[width=\textwidth]{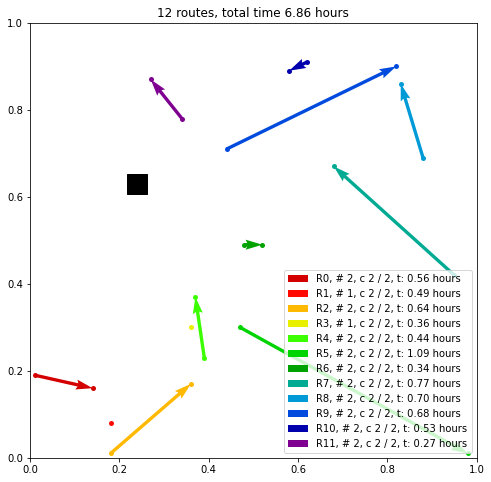}
		\caption{Non-DNN Solution} %
	\end{subfigure}
	\caption{\textbf{Social distancing policy: passenger capacity=2 people, Part II.}}\label{s21} %
\end{figure}

\subsection*{Dataset 4: 168 people and 68 houses}
\begin{figure}[H]
	\centering
	\begin{subfigure}{0.48\textwidth} %
		\includegraphics[width=\textwidth]{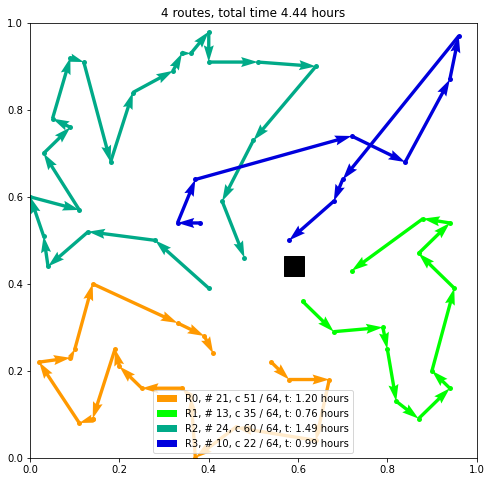}
		\caption{DNN-based Solution} %
	\end{subfigure}
	\vspace{0.1cm} %
	\begin{subfigure}{0.48\textwidth} %
		\includegraphics[width=\textwidth]{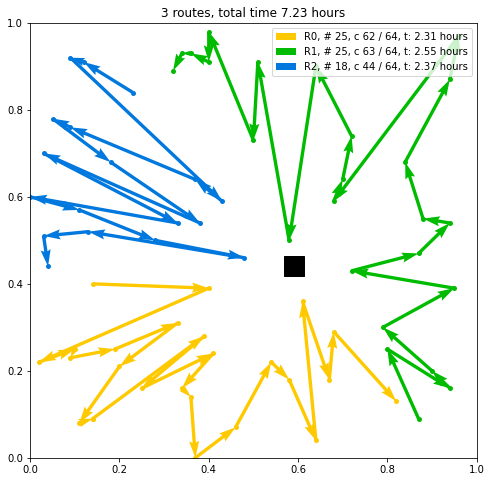}
		\caption{Non-DNN Solution} %
	\end{subfigure}
	\caption{\textbf{Social distancing policy: passenger capacity=64 people.} }\label{s22} %
\end{figure}

\begin{figure}[H]
	\centering
	\begin{subfigure}{0.48\textwidth} %
		\includegraphics[width=\textwidth]{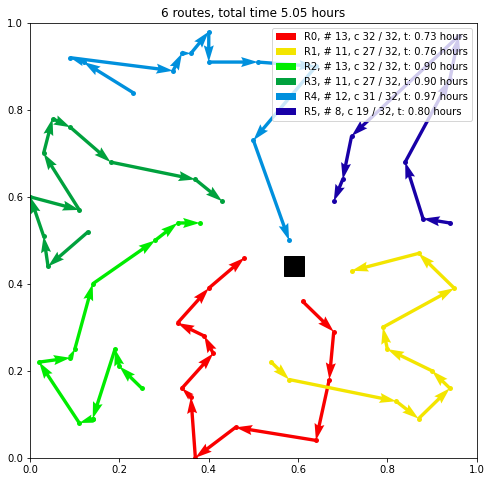}
		\caption{DNN-based Solution} %
	\end{subfigure}
	\vspace{0.1cm} %
	\begin{subfigure}{0.48\textwidth} %
		\includegraphics[width=\textwidth]{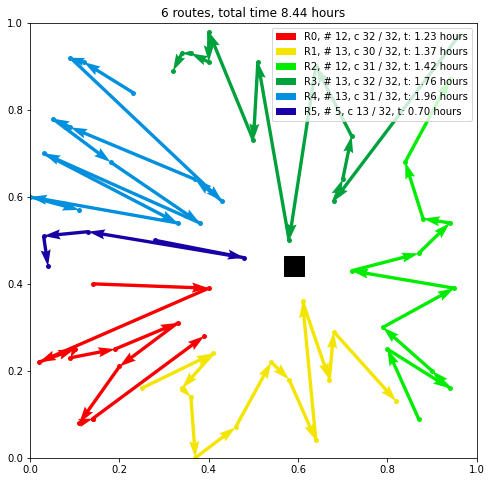}
		\caption{Non-DNN Solution} %
	\end{subfigure}
	\caption{\textbf{Social distancing policy: passenger capacity=32 people.} }\label{s23} %
\end{figure}

\begin{figure}[H]
	\centering
	\begin{subfigure}{0.48\textwidth} %
		\includegraphics[width=\textwidth]{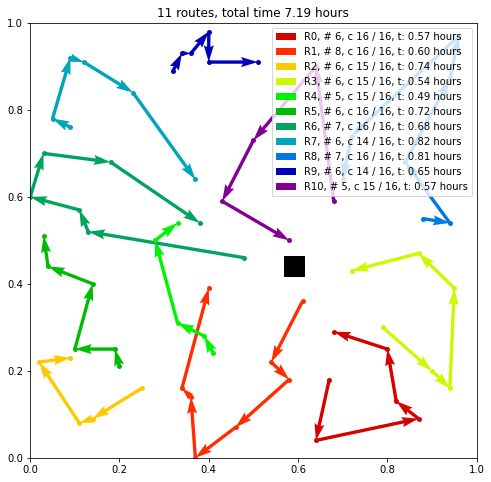}
		\caption{DNN-based Solution} %
	\end{subfigure}
	\vspace{0.1cm} %
	\begin{subfigure}{0.48\textwidth} %
		\includegraphics[width=\textwidth]{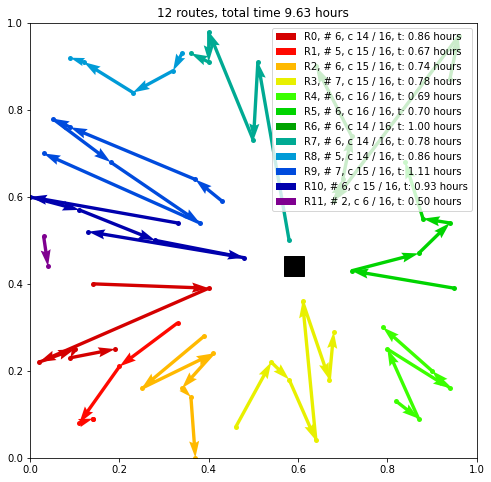}
		\caption{Non-DNN Solution} %
	\end{subfigure}
	\caption{\textbf{Social distancing policy: passenger capacity=16 people.} }\label{s24} %
\end{figure}

\begin{figure}[H]
	\centering
	\begin{subfigure}{0.48\textwidth} %
		\includegraphics[width=\textwidth]{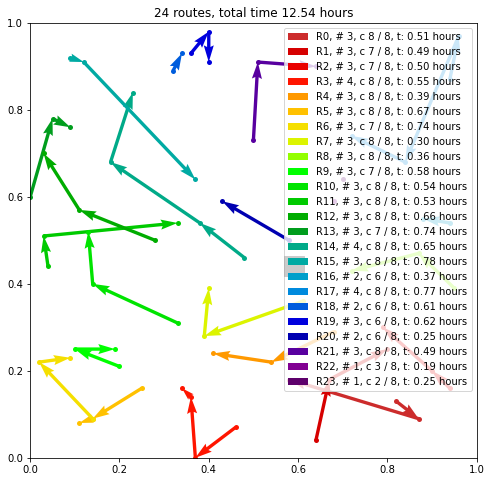}
		\caption{DNN-based Solution} %
	\end{subfigure}
	\vspace{0.1cm} %
	\begin{subfigure}{0.48\textwidth} %
		\includegraphics[width=\textwidth]{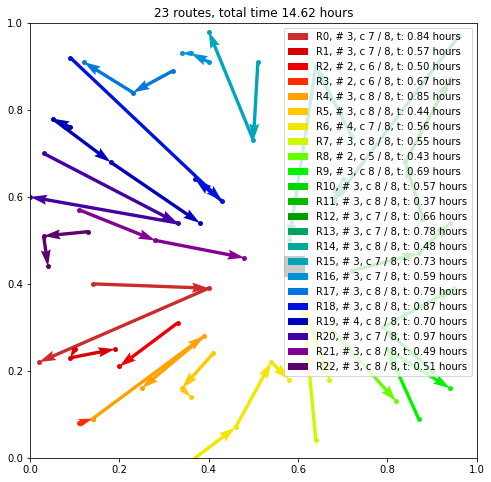}
		\caption{Non-DNN Solution} %
	\end{subfigure}
	\caption{\textbf{Social distancing policy: passenger capacity=8 people.} }\label{s25} %
\end{figure}

\begin{figure}[H]
	\centering
	\begin{subfigure}{0.48\textwidth} %
		\includegraphics[width=\textwidth]{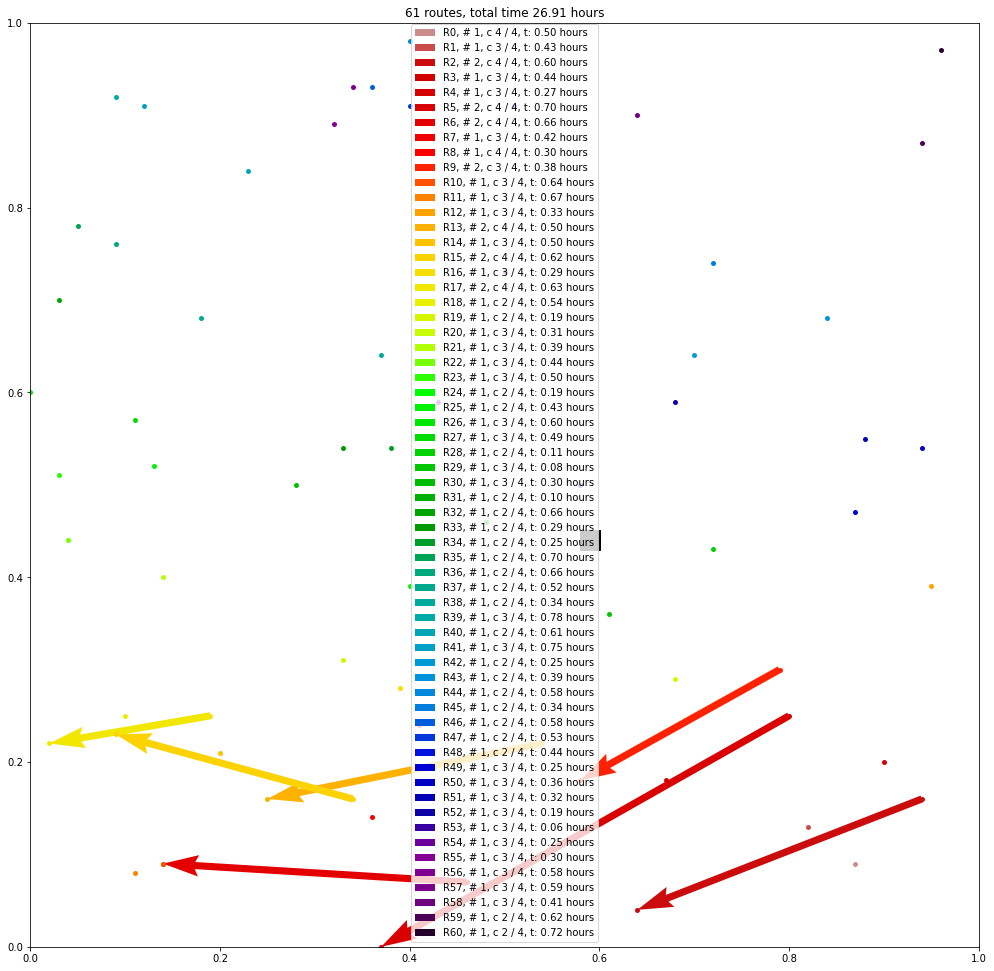}
		\caption{DNN-based Solution} %
	\end{subfigure}
	\vspace{0.1cm} %
	\begin{subfigure}{0.48\textwidth} %
		\includegraphics[width=\textwidth]{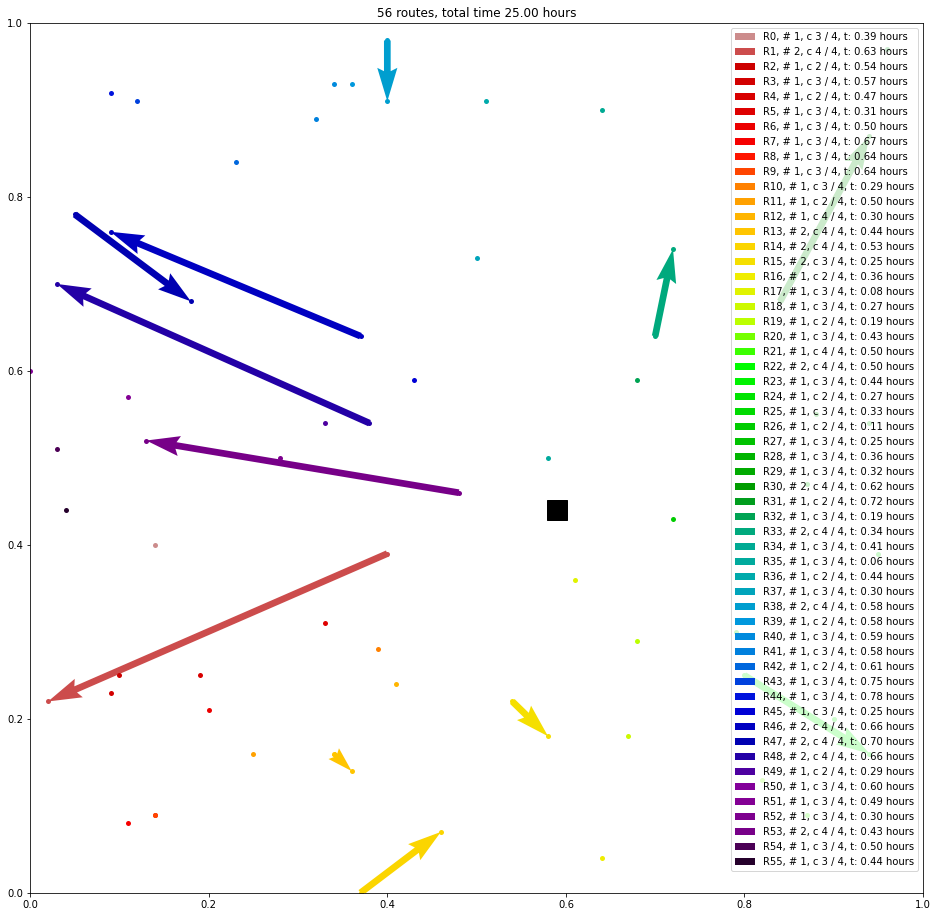}
		\caption{Non-DNN Solution} %
	\end{subfigure}
	\caption{\textbf{Social distancing policy: passenger capacity=4 people.} }\label{s26} %
\end{figure}

\begin{figure}[H]
	\centering
	\begin{subfigure}{0.48\textwidth} %
		\includegraphics[width=\textwidth]{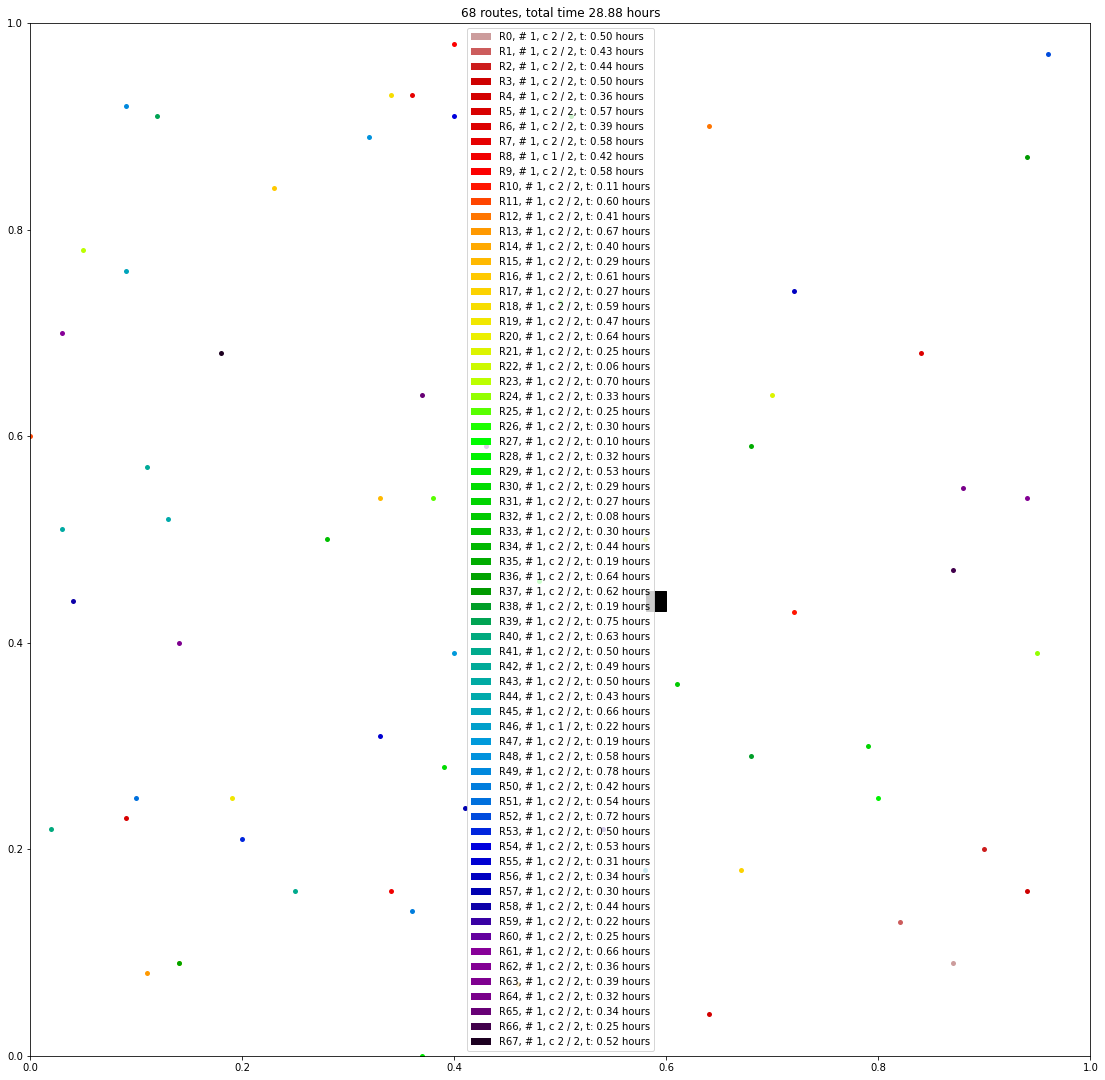}
		\caption{DNN-based Solution} %
	\end{subfigure}
	\vspace{0.1cm} %
	\begin{subfigure}{0.48\textwidth} %
		\includegraphics[width=\textwidth]{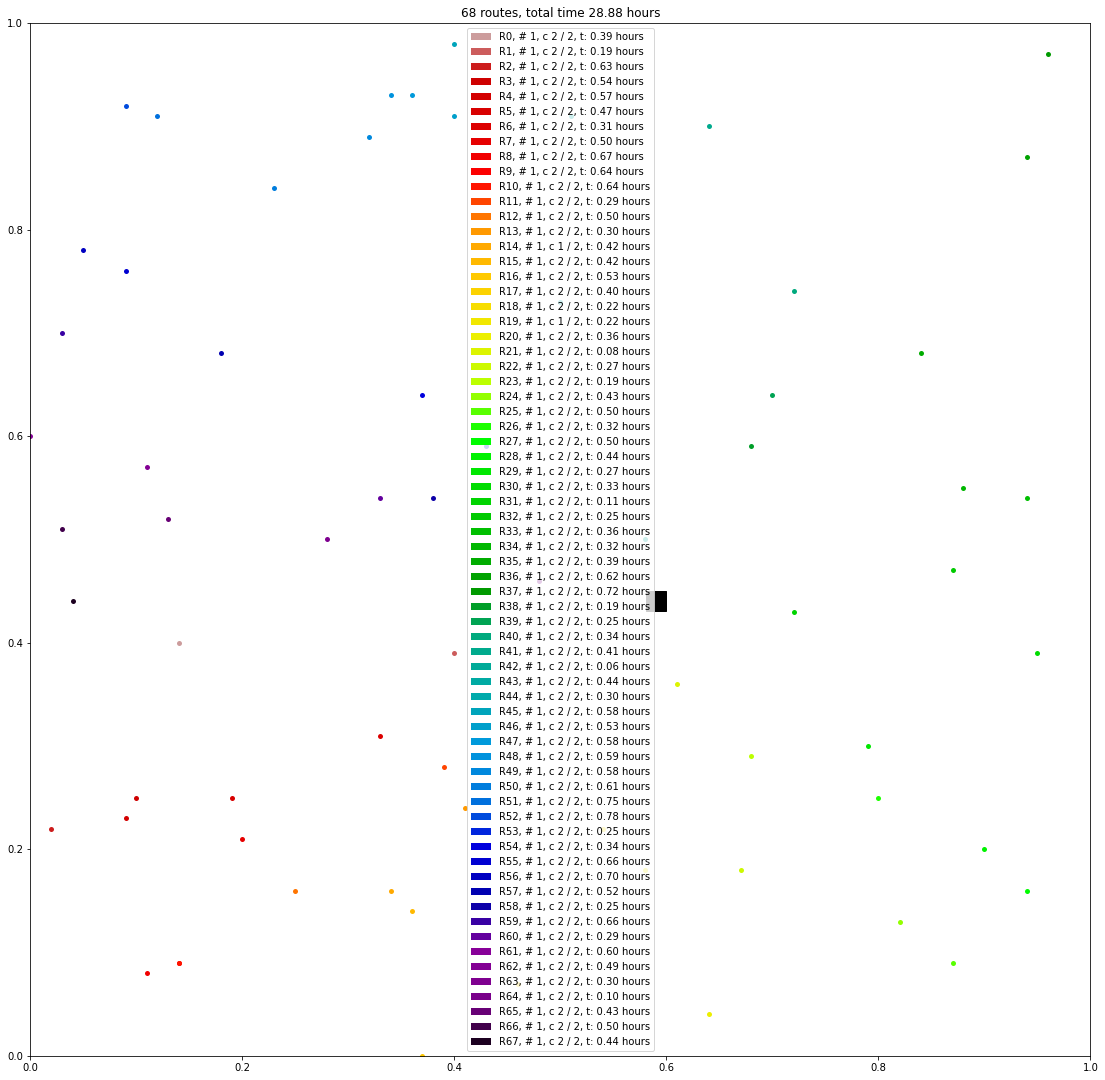}
		\caption{Non-DNN Solution} %
	\end{subfigure}
	\caption{\textbf{Social distancing policy: passenger capacity=2 people, Part I.}}\label{s27} %
\end{figure}

\begin{figure}[H]
	\centering
	\begin{subfigure}{0.48\textwidth} %
		\includegraphics[width=\textwidth]{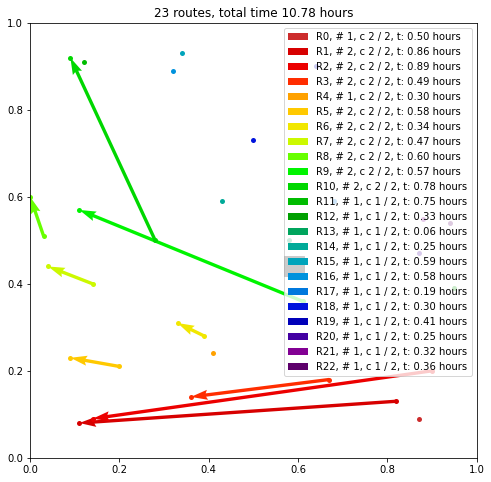}
		\caption{DNN-based Solution} %
	\end{subfigure}
	\vspace{0.1cm} %
	\begin{subfigure}{0.48\textwidth} %
		\includegraphics[width=\textwidth]{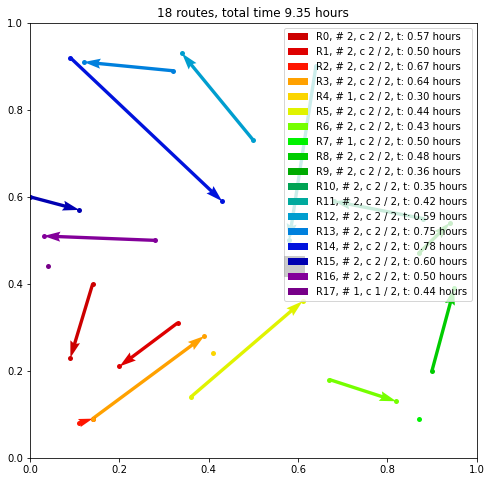}
		\caption{Non-DNN Solution} %
	\end{subfigure}
	\caption{\textbf{Social distancing policy: passenger capacity=2 people, Part II.}}\label{s28} %
\end{figure}

\begin{figure}[H]
\includegraphics[width=0.5\textwidth]{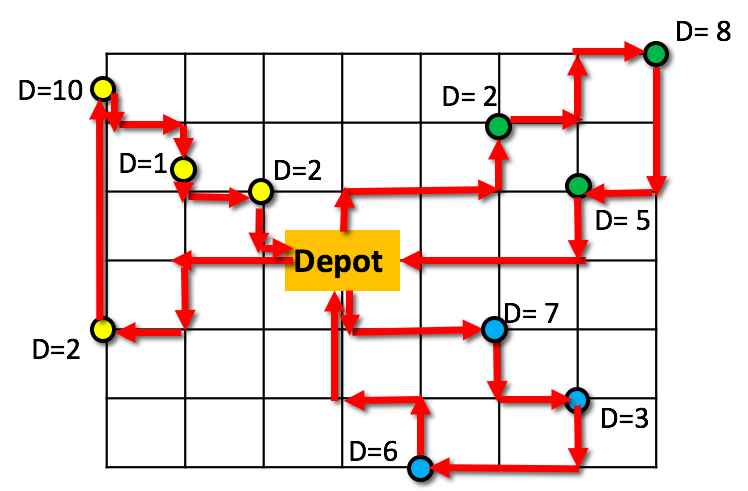}\centering
\caption{\textbf{Example plot of a solution to CVRP in Manhattan distance.} The vehicle capacity is 16 people per route.}\label{fig:cvrp_manhattan}
\centering
\end{figure}

\end{document}